\documentclass[10pt,twocolumn,letterpaper]{article}

\usepackage[pagenumbers]{cvpr} % To force page numbers, e.g. for an arXiv version

\usepackage{graphicx}
\usepackage{amsmath}
\usepackage{amssymb}
\usepackage{booktabs}

\usepackage[pagebackref,breaklinks,colorlinks]{hyperref}

\usepackage{adjustbox}
\usepackage{tabularx,booktabs}       % professional-qualit  y tables
\usepackage{amsmath}
\usepackage{graphicx}
\usepackage{mathtools}
\usepackage{caption}
\usepackage{booktabs}       % professional-quality tables
\usepackage{amsfonts}       % blackboard math symbols
\usepackage{microtype,xcolor}      % microtypography
\usepackage{wrapfig}
\usepackage{enumerate}  
\usepackage{multirow}
\usepackage{xfrac}
\newcommand{\comment}[1]{}
\usepackage[capitalize]{cleveref}
\crefname{section}{Sec.}{Secs.}
\Crefname{section}{Section}{Sections}
\Crefname{table}{Table}{Tables}
\crefname{table}{Tab.}{Tabs.}

\def\cvprPaperID{5702} % *** Enter the CVPR Paper ID here
\def\confName{CVPR}
\def\confYear{2023}

\begin{document}

\title{Warped Convolutional Networks: \\Bridge Homography to $\mathfrak{sl}(3)$ algebra by Group Convolution} 

\author{Xinrui Zhan\\
Zhejiang University\\
{\tt\small xrzhan@zju.edu.cn}
\and
Yang Li\\
East China Normal University\\
{\tt\small yli@cs.ecnu.edu.cn}
\and
Wenyu Liu\\
Zhejiang University \\
{\tt\small liuwenyu.lwy@zju.edu.cn}
\and
Jianke Zhu\\
Zhejiang University \\
{\tt\small jkzhu@zju.edu.cn}
}
\maketitle

%%%%%%%%% ABSTRACT
\begin{abstract}
		Homography has an essential relationship with the special linear group and the embedding Lie algebra structure.
		Although the Lie algebra representation is elegant, few researchers have established the connection between homography and algebra expression in neural networks. In this paper, we propose Warped Convolution Networks~(WCN) to effectively learn and represent the homography by $SL(3)$ group and $\mathfrak{sl}(3)$ algebra with group convolution. To this end, six commutative subgroups within the $SL(3)$ group are composed to form a homography. For each subgroup, a warping function is proposed to bridge the Lie algebra structure to its corresponding parameters in homography.
		By taking advantage of the warped convolution, homography learning is formulated into several simple pseudo-translation regressions. By walking along the Lie topology, our proposed WCN is able to learn the features that are invariant to homography. Moreover, it can be easily plugged into other popular CNN-based methods. Extensive experiments on the POT benchmark, S-COCO-Proj, and MNIST-Proj dataset show that our proposed method is effective for planar object tracking, homography estimation, and classification.
\end{abstract}

%%%%%%%%% BODY TEXT
\section{Introduction}

\label{sec:intro}

% \begin{figure}
% \centering
% % \setlength{\abovecaptionskip}{-0.08cm}
% \includegraphics[width=0.95 \linewidth]{fig/lie_representation_png.pdf}
%     \caption{Comparisons on semantic map construction frameworks (camera-based, LiDAR-based, Camera-LiDAR fusion methods) and our proposed LiDAR2Map that presents
%     an effective online Camera-to-LiDAR distillation scheme with a BEV feature pyramid decoder in training.} 
%     %LiDAR can achieve the promising performance with an effective BEV distillation scheme from providing camera images during training without the inference overhead.}
%     \label{fig:intro}\label{intro}
%     \vspace{-7mm}
%     % \vspace{-0.8cm}
% \end{figure}

\begin{figure}
\centering 
    \includegraphics[scale=0.35]{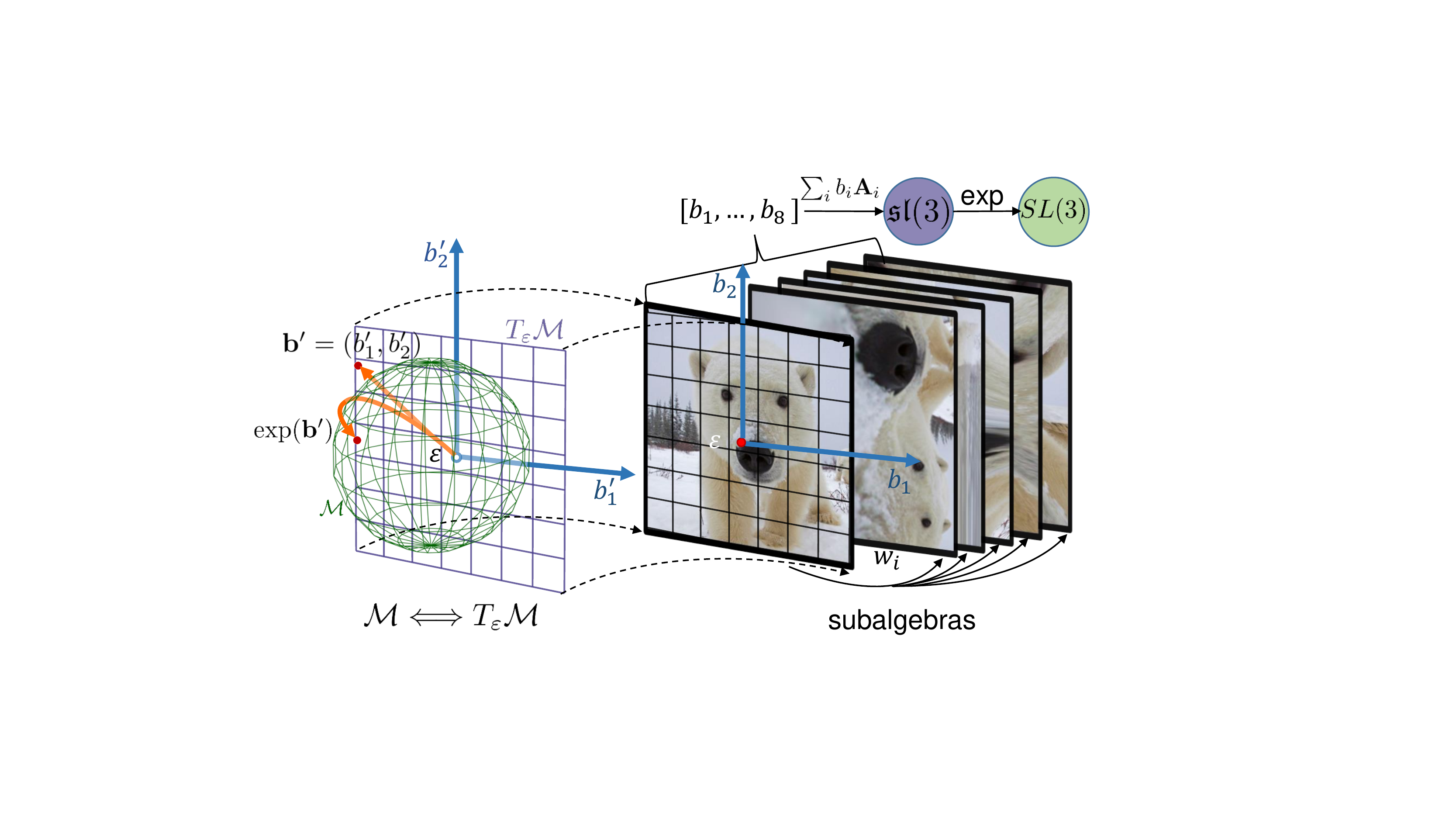}
    \vspace{-0.2in}
    \caption{A representation of the relation between a two-parameter commutative Lie subalgebra and the homography.  Left $T_\varepsilon\mathcal{M}$ (purple plane) is the tangent space of the group’s Manifold $\mathcal{M}$ (represented as a green sphere) at the identity $\varepsilon$. Their transformation is connected simply by the exponential map. $T_\varepsilon\mathcal{M}$ is a vector space. We specify it as the two-parameter Lie algebra, where the elements of its generators coefficients vector $\mathbf{b}^\prime$ are orthogonal. \comment{The movement $\mathbf{v}t$ can be represented as $\mathbf{b}^\prime= [b_1^\prime,b_2^\prime]$, where $\mathbf{v}$ is the velocity, and $t$ is the time.} The movement can be represented as $\mathbf{b}^\prime= [b_1^\prime,b_2^\prime]$. With the warp function $w$, we have each subalgebra~(right) of the $SL(3)$ satisfying the requirement, the two-parameter transformation thus becomes the translation determined by $\mathbf{b}$, where $\mathbf{b}$ denotes the generators coefficients of $SL(3)$. Then, $\mathbf{b}$ is easily connected to homography ($SL(3)$) with $\exp$ function, where A is generators for $\mathfrak{sl}(3)$.}
    % \vspace{-0.2in}
    \label{fig:lie_representation_dim}
    \vspace{-0.2in}
\end{figure}

Group convolution attracts a lot of research attention due to its underlying group structure constrain upon the learning representation. The most successful case is Convolution Neural Networks~(CNN) that is famous for the weak translation equivariance on representing visual object in image space.
Essentially, the translation equivariance is achieved due to the constraint and intrinsic topological structure of discrete groups on the image.
With a simple group structure, CNN has already been successfully and extensively used in a variety of tasks, including object detection~\cite{RFCN}, recognition~\cite{Zhou2014LearningDF}, tracking~\cite{SiamRCNN}, and alignment~\cite{Ji2020DirectionalCA}. 
To further exploit the representation capability, researchers try to extend the conventional convolution to group convolution~\cite{Macdonald2021EnablingEF,G-Conv,Weiler2018LearningSF,SteerF} with the diversity of group structures.
	
Among these group structures, special linear~($SL$) group and its embedding Lie algebra have great potential in visual representation since the corresponding homography describes the relation of two image planes for a 3D planar object with perspective transformation.
Intuitively, a laptop is always a laptop wherever you are looking from.
Every element in $SL(3)$ represents a homography of two different cameras shooting at a static 3D planar object in the scene. 
The corresponding Lie algebra space $\mathfrak{sl}(3)$ describes the changes of a camera's configuration, which means the local changes in Lie algebra coincide with the slight movement of viewpoint. Neural networks built on the space of $\mathfrak{sl}(3)$ could achieve the homography learning capability, which gives, to some extent, equivariance and invariance to the feature representation for visual objects.
This property could benefit a number of applications,
e.g. homography estimation~\cite{Japkowicz2017HomographyEF}, planar object tracking~\cite{HDN}, feature representation~\cite{STN}.

However, few researchers have investigated the relation between homography and the $\mathfrak{sl}(3)$ algebra.
Some task-oriented works~\cite{PTN, Ye2021ExploitingII,finzi2020generalizing} show the preliminary results in the application while rarely establishing the connection to the corresponding group. Benton et al.~\cite{benton2020learning} learn the invariance by parameterizing a distribution with augmentations and optimizing the training loss, simultaneously.
Carlos et al.~\cite{PTN} estimate the translation firstly and then classify the image in the log-polar coordinates~\cite{Zwicke1983ANI}, which is in fact a special case for the similarity group $Sim(2)$. 
Ye et al.~\cite{Ye2021ExploitingII} enforce a $GL(n)$-invariance property with global statistics extracted from training data, in which the gradient descent optimization should maintain the group invariance under basis changes.

Unfortunately, existing methods either are only capable of dealing with several subgroups of $SL(3)$ and their corresponding transformations, or purely enforce the equivariance learning by tremendous data augmentation on the image domain. Dehmamy et al.~\cite{LConv} introduce L-conv method for unknown groups to learn Lie algebra in an unsupervised fashion.
Macdonald et al.~\cite{Macdonald2021EnablingEF} sample multiple times~(32$\times$) from Haar measure and achieve the SL(3) equivariant networks. 
Both methods have drawbacks due to either the inferior performance or heavy computational cost.

Our goal is to connect the representation learning of homography with $\mathfrak{sl}(3)$ algebra for neural networks in an efficient way. 
When the representation is based on Lie algebra, we could investigate the potential in the algebra space with its mathematical property. 
For instance, the feature representation is consistent with human's intuitive perception, as the transformation walks along the geodesic curve in algebra space.
This allows the networks to have very robust feature representation for different transformations and have the capability to neglect the noise in training. 
Additionally, this is helpful for learning the implicit transformation from a single image, which may facilitate the applications such as congealing~\cite{LearnedMiller2006DataDI} and facade segmentation~\cite{DeepFacade }. 

Inspired by the warped convolution~\cite{Henriques2017WarpedCE}, we construct six commutative subgroups within the $SL(3)$ group from the Lie algebra $\mathfrak{sl}(3)$ generators to learn homography.
For each subgroup, a warping function is proposed to bridge the Lie algebra structure to its corresponding parameters in homography.
As the constructed subgroups are Abelian groups, the group convolution operation can be formulated into conventional convolution with a well-designed warping function.
In this paper, we propose Warped Convolution Networks~(WCN) to compose these subgroups to form an $SL(3)$ group convolution by predicting several pseudo-translation transformations. 
Our proposed WCN is able to handle noncommutative groups and learn the invariant features for homography.
The main contribution can be summarized as follows:
\begin{itemize}
    \setlength{\parskip}{0pt}	

    \item A general neural network for efficient Lie group convolution. Our proposed framework can deal with most of the Lie groups by easily combining different basis, which is able to learn the invariant features for different neural networks.
    
    \item A robust homography estimator based on WCN. To the best of our knowledge, it is the first work to directly estimate homography along with the $SL(3)$ group and its algebra topology. 

    \item Extensive experimental evaluation of three datasets demonstrates that our approach is effective for three computer vision tasks and offers more potentials for robustly learning the large and implicit transformations.
\end{itemize}
\begin{figure*}[htbp]
    \centering 
    \includegraphics[scale=0.36]{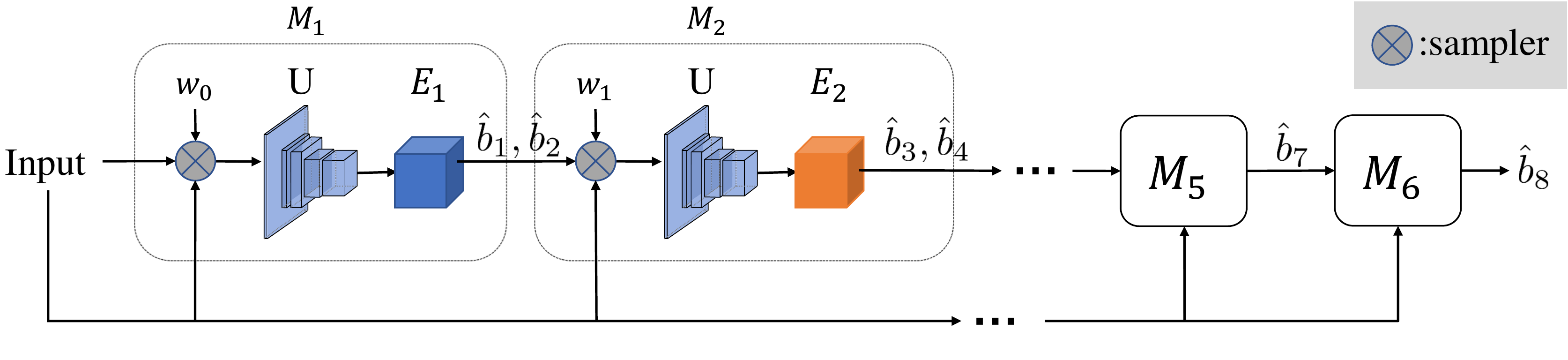}
    \vspace{-0.15in}
    \caption{ Overview of WCN. $M_1$ is a basic module of WCN, and $w_i$ is the $i_{th}$ warp function. The output(s) is the transformed version of the input image(s). The total number of inputs depends on the tasks. $U$ is a CNN backbone, and $E_1$ is a translation estimator. The sampler is used to re-sample the input according to $w$ and the previously estimated parameters. We do not specify the detailed structure for its generality, however, details for specific tasks can be found in the Experiments and Appendix.}
    \label{fig:pipeline}
    \vspace{-0.2in}
\end{figure*}

%-------------------------------------------------------------------------
\vspace{-2mm}
\section{Related Work}

In this section, we discuss the related prior studies, including the equivariant networks and transformation learning. 
% 	Many related works are limited to the finite groups~\cite{} such as p4~\cite{G-Conv} and E(2)~\cite{weiler2019general}. 
Cohen et al.~\cite{G-Conv} present the fundamental work on equivariance of CNNs representations for the image transformations, where the underlying properties of symmetry groups~\cite{cohen2015transformation} are investigated. They replace the translational convolutions with group convolutions and propose Group equivariant convolutional networks (G-CNNs). For the continuous groups, Cohen et al.~\cite{G-Conv} discretize the group and employ the harmonic function as the irreducible representation of convolution~\cite{weiler20183d, Weiler2018LearningSF, SteerF}. Recently, Macdonald et al.~\cite{Macdonald2021EnablingEF} change the convolution formula and modify the different layers of CNNs. They use the Schur-Poincaré formula for the derivative of exponential map, which enables sampling from Haar measure for arbitrary Lie groups. Dehmamy et al.~\cite{LConv} propose an unsupervised learning approach to automatically learn the symmetry based on Lie algebra in CNNs. 
It performs inferior to the baseline when the symmetric group is already known. 
All these methods introduce the various architectures that are evidently different from the original convolutional layer and are difficult to directly make use of the popular CNN backbones. Moreover, they can only deal with the classification problem.   

Early work learns the transformation representation by an auto-encoder~\cite{Hinton2011TransformingA}. It attempts to build a generative model, where the target is a transformed input image. Lin et al.~\cite{Lin2021AutoEncodingTI} change the parameterization and project the distance onto $SO(3)$. STN~\cite{STN} introduces a spatial transformation network to manipulate the data in the network without supervision on the transformation. All these methods have
difficulty in estimating the transformations since the networks can only inference once for guessing and the parameters are entangled and highly coupled.
ESM~\cite{ESM} parameterizes the arguments as the Lie algebra basis to estimate the $SL(3)$ group.
However, their parameters lose the interpretability in an image transformation.
Henriques et al.~\cite{Henriques2017WarpedCE} employ the warp function on convolution and implement two-parameter group equivariance, since there are possibly utmost two independent dimensions in an image. HDN~\cite{HDN} recently decomposes the homography into two groups and estimates them in order, which loses the equivariance for the residual parameters. Besides, it requires an additional homography estimator to find the corners. 
Recently, deep learning-based approaches predict the homography mainly by estimating the corner offsets~\cite{Nguyen2018UnsupervisedDH, Zhang2020ContentAwareUD} or pixel flows~\cite{Zeng2018RethinkingPH}. They focus on the local movements in the image space, which are incapable of estimating the large transformation. Our proposed approach can be viewed as a general case of the warped convolution in 2D space, which is able to handle the most sophisticated 2D Lie group $SL(3)$. 
% \textcolor{red}{
It is noteworthy that 
% compared to the proposed method, 
HDN~\cite{HDN} also estimates parameters from two groups based on warp functions~\cite{Henriques2017WarpedCE}.
However, 
% they borrow the concept similarly from the \cite{Henriques2017WarpedCE} where they only explain the theory based on the group and 
% HDN does not bring any new warp functions for the other 4 parameters.
they only employ the rotation-and-scale subgroup and
refine the transformation from a corner regression-based homography estimator.  
Differently, our proposed method bridges the gap from the similarity group and two-parameter group to any subgroup of the $SL(3)$ and completes a full homography based on group convolution.

%-------------------------------------------------------------------------
\vspace{-2mm}
\section{Method}\label{sec:method}

The main objective of this work is to formulate a full homography on Lie subalgebras with several equivariant warped convolutions for 2D projective transformation. 
Since the warped convolution only implements two-parameter equivariance, a possible way is to combine the several warped convolutions. 
In general, the 2D projective transformation is an $SL(3,\mathbb{R})$ group having a few subgroups. 
Our proposed method divides this group into several one or two-parameter subgroups, whose Lie algebras are the subalgebras of $\mathfrak{sl}(3)$. 
As explained in the Fig~\ref{fig:lie_representation_dim}, the warped convolution can be employed to achieve the equivariance for each single or two-parameter transformation. Finally, they are combined to obtain the full transformation. In this section, we first introduce the fundamental of the warped convolution, and then describe our proposed method.

% The main purpose of this work is to formulate a full homography on Lie subalgebras with several equivariant warped convolutions for 2D projective transformation.
% Since the warped convolution only implements two-parameter equivariance, a possible way is to combine the several warped convolutions. 
% In general, the 2D projective transformation is an $SL(3,\mathbb{R})$ group having a few subgroups. 
% Our proposed method divides this group into several one or two-parameter subgroups, whose Lie algebras are the subalgebras of $\mathfrak{sl}(3)$. 
% As a result, the warped convolution can be employed to achieve the equivariance for each single or two-parameter transformation. Finally, they are combined to obtain the full transformation. In this section, we first introduce the fundamental of the warped convolution, and then describe our proposed method.

\subsection{Warped Convolution}\label{sec:warped_convolution}
The key to CNNs' equivariance is their convolution layers, in which the basic operation is the convolution of an image $I\in \mathbb{R}^{n\times n}$ and a convolution kernel $F\in \mathbb{R}^{n^\prime\times n^\prime}$. 
By employing the Dirac delta function on the image and kernel~\cite{Henriques2017WarpedCE}, the convolution formula can be treated as a special case of continuous function as follows,
\vspace{-2mm}
\begin{equation}
(I*F)(\mathbf{v}) = \int I(\mathbf{v}+\mathbf{u}) F(-\mathbf{u}) d\mathbf{u}.
\label{eq:cont-conv}
\end{equation}
\vspace{-1mm}
where $\mathbf{u}$ and $\mathbf{v}$ are the coordinates for $I$ and $F$.
For the sake of convenience, we shift the image $I$ instead of $F$ in the convolution equations. 
To prove the equivariance, we define the transformation operator as $\pi_l: \mathbf{u} \longmapsto	\mathbf{u}+\mathbf{l}$. Hence, the equivariance concerning the translation $\mathbf{l}$ can be easily proved as: $(\pi_l(I)*F)(\mathbf{v}) = \int I(\mathbf{u}+(\mathbf{v}+\mathbf{l})) F(-\mathbf{u}) d\mathbf{u} = (\pi_l(I*F))(\mathbf{v})$.

The standard convolution only takes into account the translation equivariance in the image. 
For the equivariance of other groups in the image domain, Henriques and Vedaldi~\cite{Henriques2017WarpedCE} suggest an intuitive solution that defines a function of a group action on the image as below,
\vspace{-1mm}
\begin{equation}
(\tilde{g}*\tilde{h})(q) = \int_G g(pqx_0)h(p^{-1}x_0)d\xi(p).
\label{eq:group-conv}
\end{equation}
% \vspace{-1mm}
Eq.~\ref{eq:group-conv} provides the convolution of two real functions $\tilde{g}$ and $\tilde{h}$. $g(px_0)=\tilde{g}$ and $h(px_0)=\tilde{h}$ are defined on a subset $\Omega \in \mathbb R^2$, where $x_0\in \Omega$ is an arbitrary constant pivot point.
Compared to the convolution on the image, the operation $\mathbf{u}+\mathbf{v}$ becomes $pq$ for the function $g$ and $h$, where $p,q\in G$. 
The integration is under the Haar measure $\xi$, which is the only measure invariant to the group transformation.
Since Eq.~\ref{eq:group-conv} is still defined over the group, it needs to be further simplified. 
As illustrated in the Fig~\ref{fig:lie_representation_dim}, a simple approach is to define $G$ as a Lie group, which can be projected onto the Lie algebra $\mathfrak{m}$. $\mathfrak{m}$ is a vector space tangent at identity  $\boldsymbol{\varepsilon}$ of the group manifold $\mathcal{M}$, whose base coefficients $\mathbf{b}$ are easy to map to the Cartesian vector space $V \in \mathbb{R}^m$. The dimension $m$ is the degrees of freedom for $\mathcal{M}$.
This mapping allows us to estimate the Lie algebra on the real plane $\mathbb{R}^2$. 
% 	Moreover, we can utilize the property that is not explored in~\cite{Henriques2017WarpedCE}. 
Once the element of Lie algebra is obtained, it could be mapped to $\mathcal{M}$. 
Therefore, an exponential map $\exp:V\to G$ is employed to connect the Cartesian vector space with the $\mathcal{M}$, where $V$ is a subset of $\mathbb{R}^2$. 
We therefore have the warped image $g_w(\mathbf{u}) = g(\exp(\mathbf{u})x_0)$. 
%\begin{equation}T=
%	g_w(p) = I(exp(p)x_0) 
%\label{eq:warp-func}
%\end{equation}
Thus, Eq.~\ref{eq:group-conv} can be rewritten as:
\vspace{-4mm}
\begin{equation} 
(\tilde{g}*\tilde{h})(\exp(\mathbf{v})) = \int_V g_w(\mathbf{u}+\mathbf{v}) h_w(-\mathbf{u})d\mathbf{u}.
\label{eq:warped-conv}
\end{equation}  
% \vspace{-2mm}
where $h_w(\mathbf{u})=h(\exp(\mathbf{u})x_0)$.
Obviously, Eq.~\ref{eq:warped-conv} has the same formulation as Eq.~\ref{eq:cont-conv}. This achieves the equivariance to the transformation belonging to the Lie group by performing a conventional convolution after warping the image which connects the warped convolution and group convolution.
% \vspace{-2mm}
\subsection{Warped Convolutional Networks}
% \vspace{-1mm}
%The projective transformation (8 parameters) corresponds to the rigid transformation (6 DoFs) in 3D space. 
As introduced in Section ~\ref{sec:warped_convolution}, warp function is used for implementing the estimation from the Lie algebra, which shares the equivariance and properties of Lie algebra. 
However, in a warped image $g_w$, one can only estimate at most two independent Lie algebra parameters for its dimensional restriction. To accomplish the goal for estimation purely from the Lie algebra for $\mathfrak{sl}(3)$,
we thus employ the compositional method to estimate the Lie subalgebras and combine the subgroups in order. 
To warp the image by a series of functions, the $SL(3)$ generators need to be defined before the composition. A generator of the Lie algebra is also called the infinitesimal generator, which is an element of the Lie algebra.
In this paper, we choose the widely used 2D projective group decomposition~\cite{Harltey2003MultipleVG}, whose corresponding eight generators of its subgroups are defined as follows,
% eight-parameters Lie algebra is used to compose the Lie algebra $\mathfrak{s}\mathfrak{l}(3)$. 
% Specifically, these generators are defined as follows,
% \begin{shrinkeq}{-2ex}
% \vspace{-1mm}
\begin{small}
    % \vspace{-0.1in}
    \setlength{\arraycolsep}{0.7pt}
    \begin{align}
    \nonumber
    %\scriptsize{
    %			\mathbf{H}(\mathbf{x})=& 
    %			\mathbf{H_t}\cdot\mathbf{H_s}\cdot \mathbf{H_{sc}}\cdot\mathbf{H_{sh}}\cdot\mathbf{H_{p1}}\cdot\mathbf{H_{p2}}\\	\nonumber=&
    % \mathclap
    % \begin{scriptsize}
    \hspace{-1.0em}\mathbf{A_1}\hspace{-0.45em} =&\hspace{-0.45em} 
    % \mathbf{A_1} =&
    %			\begin{bmatrix}
    \begin{bmatrix}
    0 &0 & 1 \\
    0 &0 & 0 \\
    0 & 0 & 0 
    \end{bmatrix}
    \hspace{-0.1em}\mathbf{A_2}\hspace{-0.45em} =\hspace{-0.45em} 
    % \mathbf{A_2} = 
    \begin{bmatrix}
    0 &0 & 0 \\
    0 &0 & 1 \\
    0 & 0 & 0 
    \end{bmatrix}
    \hspace{-0.1em}\mathbf{A_3}\hspace{-0.45em} = \hspace{-0.45em}
    % \mathbf{A_3} = 
    \begin{bmatrix}
    0 &-1 & 0 \\
    1 &0 & 0 \\
    0 & 0 & 0	
    \end{bmatrix}
    \hspace{-0.1em}\mathbf{A_4}\hspace{-0.45em} = \hspace{-0.45em}
    % \mathbf{A_4} = 
    \begin{bmatrix}
    0 &0 & 0 \\
    0 &0 & 0 \\
    0 & 0 & -1	
    \end{bmatrix} \\
    \hspace{-0.1em}\mathbf{A_5}\hspace{-0.45em}  =& \hspace{-0.45em}
    % \mathbf{A_5}  =& 
    \begin{bmatrix}
    1 &0 & 0 \\
    0 &-1 & 0 \\
    0 & 0 & 0	
    \end{bmatrix}
    \hspace{-0.1em}\mathbf{A_6}\hspace{-0.45em} = \hspace{-0.45em}
    % \mathbf{A_6} = 
    \begin{bmatrix}
    0 &1 & 0 \\
    0 &0 & 0 \\
    0 & 0 & 0	
    \end{bmatrix}
    \hspace{-0.1em}\mathbf{A_7} \hspace{-0.45em}=\hspace{-0.45em} 
    % \mathbf{A_7} =
    \begin{bmatrix}
    0 &0 & 0 \\
    0 &0 & 0 \\
    1 & 0 & 0	
    \end{bmatrix}
    \hspace{-0.1em}\mathbf{A_8}\hspace{-0.45em} = \hspace{-0.45em}
    % \mathbf{A_8} = 
    \begin{bmatrix}
    0 &0 & 0 \\
    0 &0 & 0 \\
    0 & 1 & 0	
    \end{bmatrix}
    % \end{scriptsize}
    \label{eq:decompose_matrix}
    \end{align}
\end{small}
\vspace{-1mm}

For each $i_{th}$ generator $\mathbf{A}_i$, we construct a one-parameter group. The other dimension for $g_w$ could be viewed as an identity transformation group, which is commutative to the one-parameter group. 
As a result, the equivariance is also valid in the case of one-parameter group. 
We choose the generators and compose them corresponding to two or one parameter group for warping, as long as they are commutative. In this paper, we propose to compose the generators of  $\mathfrak{sl}(3)$ into six Lie subalgebras as [$b_1\mathbf{A_1}+ b_2\mathbf{A_2}, b_3\mathbf{A_3}+b_4\mathbf{A_4}, b_5\mathbf{A_5}, b_6\mathbf{A_6}, b_7\mathbf{A_7}, b_8\mathbf{A_8}$], where $[b_1,b_2,...,b_8]$ are the elements of the generator coefficients vector $\mathbf{b}$. (More details can be found in Appendix B)
% ~\ref{sec:implementation_details}.) 
For better symbol presentation, we re-parameterize $\mathbf{b}$ into a homography-friendly format to link the Lie algebra with the homography $\mathbf{H}$. The resulting intermediate variables vector $\mathbf{x} = [t_1,t_2,\theta, \gamma, k_1, k_2, \nu_1, \nu_2] = [b_1, b_2,b_3, \exp(b_4), \exp(b_5), b_6, b_7, b_8 ]$. We will introduce the definition of $\mathbf{x}$ in the next sub-section. 
Therefore, the six Lie subalgebras corresponding subgroups~($\mathbf{H}_t$, $\mathbf{H}_s$,$\mathbf{H}_{sc}$, $\mathbf{H}_{sh}$, $\mathbf{H}_{p1}$, $\mathbf{H}_{p2}$) parameterized by $\mathbf{b}$ are defined as follows,
% \begin{small}
%     \setlength{\arraycolsep}{1.2pt}
%     \begin{align}
%     %\nonumber
%     %\scriptsize{
%     \mathbf{H}(\mathbf{x})=& 
%     \mathbf{H_t}\cdot\mathbf{H_s}\cdot \mathbf{H_{sc}}\cdot\mathbf{H_{sh}}\cdot\mathbf{H_{p1}}\cdot\mathbf{H_{p2}} \\	\nonumber=&
%     \begin{bmatrix}
%     1\, &0 & t_1 \\
%     0\, &1 & t_2 \\
%     0 & 0 & 1 
%     \end{bmatrix}
%     \begin{bmatrix}
%     \gamma\, \cos\theta & -\gamma\, \sin\theta & 0 \\
%     \gamma\, \sin\theta & \gamma\, \cos\theta & 0 \\
%     0 & 0 & 1 
%     \end{bmatrix}
%     \begin{bmatrix}
%     k_1 & 0 & 0 \\
%     0 & 1/k_1 & 0\\
%     0 & 0 & 1 
%     \end{bmatrix}\\
%     \nonumber&
%     \begin{bmatrix}
%     1 & k_2 & 0 \\
%     0 & 1 & 0\\
%     0 &  0 & 1 
%     \end{bmatrix}
%     \begin{bmatrix}
%     1 & 0 & 0 \\
%     0 & 1 & 0\\
%     \nu_1 &  0 & 1 
%     \end{bmatrix}
%     \begin{bmatrix}
%     1 & 0 & 0 \\
%     0 & 1 & 0\\
%     0 & \nu_2 & 1 
%     \end{bmatrix}
%     %	= \mathbf{H}^S \cdot \mathbf{ H}^{\Lambda}
%     %}
%     \label{eq:decompose_matrix}
%     \end{align}
% \end{small}
\begin{small}
    \setlength{\arraycolsep}{1.0pt}
    \begin{align}
    %\nonumber
    %\scriptsize{
    \mathbf{H}(\mathbf{x})=& 
    \mathbf{H_t}\cdot\mathbf{H_s}\cdot \mathbf{H_{sc}}\cdot\mathbf{H_{sh}}\cdot\mathbf{H_{p1}}\cdot\mathbf{H_{p2}} \\	\nonumber=&
    \begin{bmatrix}
    1\, &0 & b_1 \\
    0\, &1 & b_2 \\
    0 & 0 & 1 
    \end{bmatrix}
    \begin{bmatrix}
    \exp(b_4) \cos(b_3) & -\exp(b_4) \sin(b_4) & 0 \\
    \exp(b_4) \sin(b_3) & \exp(b_4) \cos(b_4) & 0 \\
    0 & 0 & 1 
    \end{bmatrix}\\
    \nonumber&
    \begin{bmatrix}
    \exp(b_5) & 0 & 0 \\
    0 & \exp(-b_5) & 0\\
    0 & 0 & 1 
    \end{bmatrix}
    \begin{bmatrix}
    1 & b_6 & 0 \\
    0 & 1 & 0\\
    0 &  0 & 1 
    \end{bmatrix}
    \begin{bmatrix}
    1 & 0 & 0 \\
    0 & 1 & 0\\
    b_7 &  0 & 1 
    \end{bmatrix}
    \begin{bmatrix}
    1 & 0 & 0 \\
    0 & 1 & 0\\
    0 & b_8 & 1 
    \end{bmatrix}.
    %	= \mathbf{H}^S \cdot \mathbf{ H}^{\Lambda}
    %}
    \label{eq:decompose_matrix}
    \end{align}
\end{small}

Based on the above equation, we propose Warped Convolutional Networks~(WCN) to learn homography by six modules $[M_1,...,M_6]$ as depicted in the Fig.~\ref{fig:pipeline}. 
Each module $M_i$ has three components, a shared backbone $U$, a translation estimator $E_i$ and a warp function $w_i$. $w_0$ is the identity warping. The translation does not need a warp function since the original image offsets already denote the offsets in the corresponding algebra. 
% 	Since the warped convolution in Eq.~\ref{eq:warped-conv} has the same formulation as the ordinary convolution, 
According to Eq.~\ref{eq:warped-conv},
recovering the Lie algebra parameters is equivalent to estimating a pseudo-translation in the subgroup to which the warp function transfers the image space.
For each module, the input image is resampled with a specially designed warp function $w_i$, fed to the backbone $U$ and estimator $E_i$ to obtain a pseudo-translation in the corresponding subalgebra.  
Note that we predict $\mathbf{b}$ essentially, and $\mathbf{x}$ is just a function of $\mathbf{b}$ for a convenient expression. 
% That is the essential difference between the proposed method and other methods, meanwhile, directly estimating the $\mathbf{x}$ is infeasible for neural networks 
% parameterization. 
% 	More specifically, each input is fed to the backbone $U$ to obtain the deep feature, then the pseudo-translation is estimated in $E_i$. 
% 	According to $w_i$, we resample the warped image, and finally pass to the next similar modules in turns.  
$E_i$ is different for each module to adapt to the different subalgebras. 
Finally, we obtain the output $\mathbf{x}$ and compose them to the transformation matrix as Eq.~\ref{eq:decompose_matrix}.
% 	However, there exist other parameters that affect the warped image in addition to each group. They share the same $U$ each time.
% 	However, besides the parameters consist each group, the other parameters of $b$ also affect the warped image. 
Theoretically, the parameters of the proposed six groups may affect each other in the warped image domain. Thus, the groups must be estimated in a cascade fashion. 
Fortunately, we found that the networks localize the object's position very well in most vision tasks, even with large deformations or distortions. 
We argue that the networks manage to learn the invariant feature for the target to compensate for the interdependence. 
Intuitively, we transfer all 8 parameters of Lie algebra $\mathfrak{sl}(3)$ into 6 subalgebras that can be solved by pseudo-translation estimation. In the warped image domain, the pseudo-translation is more significant in contrast to other transformations. Thereby, we take advantage of this property to estimate the subalgebra in each warped image.
Please refer to Appendix for more details.
% More details can be found in the Appendix. 
% \vspace{-4mm}

\subsection{Warp Functions}
As illustrated in Eq.~\ref{eq:warped-conv}, 
%  	we confirm the kernel of implementing one or two-parameter equivalence is to find a proper warp function.
the key to recovering one or two-parameter transformation is to find a proper warp function so that the pseudo-translation shift in the warped image is equivalent to the linear changes of element on the corresponding Lie algebra. 
To this end, we define the warp function as $w(\mathbf{b}^\prime) = \mathbf{u^\prime} $,
%  	Concretely, it is to construct a function of action on the image about the image coordinates $\mathbf{u}= (u_1, u_2)$, satisfying the following equation from the original input image coordinate $\mathbf{u}$ to warped image coordinate$(u_1, u_2)$:
% 	\begin{equation}
% 	w(\mathbf{x}^\prime, \mathbf{u}) = (u_1^\prime, u_2^\prime)^T = \mathbf{H}(\mathbf{x}^\prime)\mathbf{u}^T
% 	\label{eq:warp-constriant}
% 	\end{equation}
where $\mathbf{b}^\prime=(b_1^\prime, b_2^\prime)$ is the specific two-parameter coefficient vector of the warp function for two-parameter Abelian group. For one-parameter group, $b_2^\prime$ is the identity Lie algebra parameter ($b_\epsilon$) .
% 	If there is only one parameter, the other parameter in $\mathbf{x}^\prime$ is the original $u_1$ or $u_2$. 
% 		We use $\mathbf{u}$ to represent the point coordinates in the image $I$,  $\mathbf{\mu}=(\mu_1, \mu_2)$ is adopted to denote the point coordinate in the warped image $g_w$, and 
$\mathbf{u^\prime}=(u_1^\prime, u_2^\prime) $ denotes the re-sampled point in the transformed $I$.  $\mathbf{\mu}=(\mu_1, \mu_2)$ is adopted to denote the point coordinate in the warped image $g_w$. The proof can be found in the Appendix E. 
% 	\begin{equation}
% 	\left\{
% 	\begin{aligned}
% 	&\mu_1 =  \Psi(\Phi(u_1))\\
% 	&\mu_2 =  \Psi(\Phi(u_2))
% 	\end{aligned}
% 	\right.
% 	\label{eq:warp-constriant}
% 	\end{equation}
%where the $\phi$ denotes the warped image coordinates, 
%it is wrong
% 	where $\Phi(u)$ is the  inverse function of $\mathbf{H}\cdot \mathbf{u}$ to the two element of Lie algebra. $\Psi$ is a linear function. Thereby, the final warped coordinate movement is the linear transformation of the element in the Lie algebra. The condition is that $\Phi(u)$ is a bijective function.  
% 	In Eq.~\ref{eq:grid_lp} and other sampling functions, 
% 	We use $\mathbf{u}$ to represent the point coordinates in the image $I$,  $\mathbf{\mu}=(\mu_1, \mu_2)$ is adpoted to denote the point coordinate in the warped image $g_w$, and $\mathbf{u^\prime}=(\mu_1^\prime, \mu_2^\prime) $ denotes the resampled point in $I$ when warping. 

\begin{figure}
    % \centering 
    \includegraphics[width=1\linewidth]{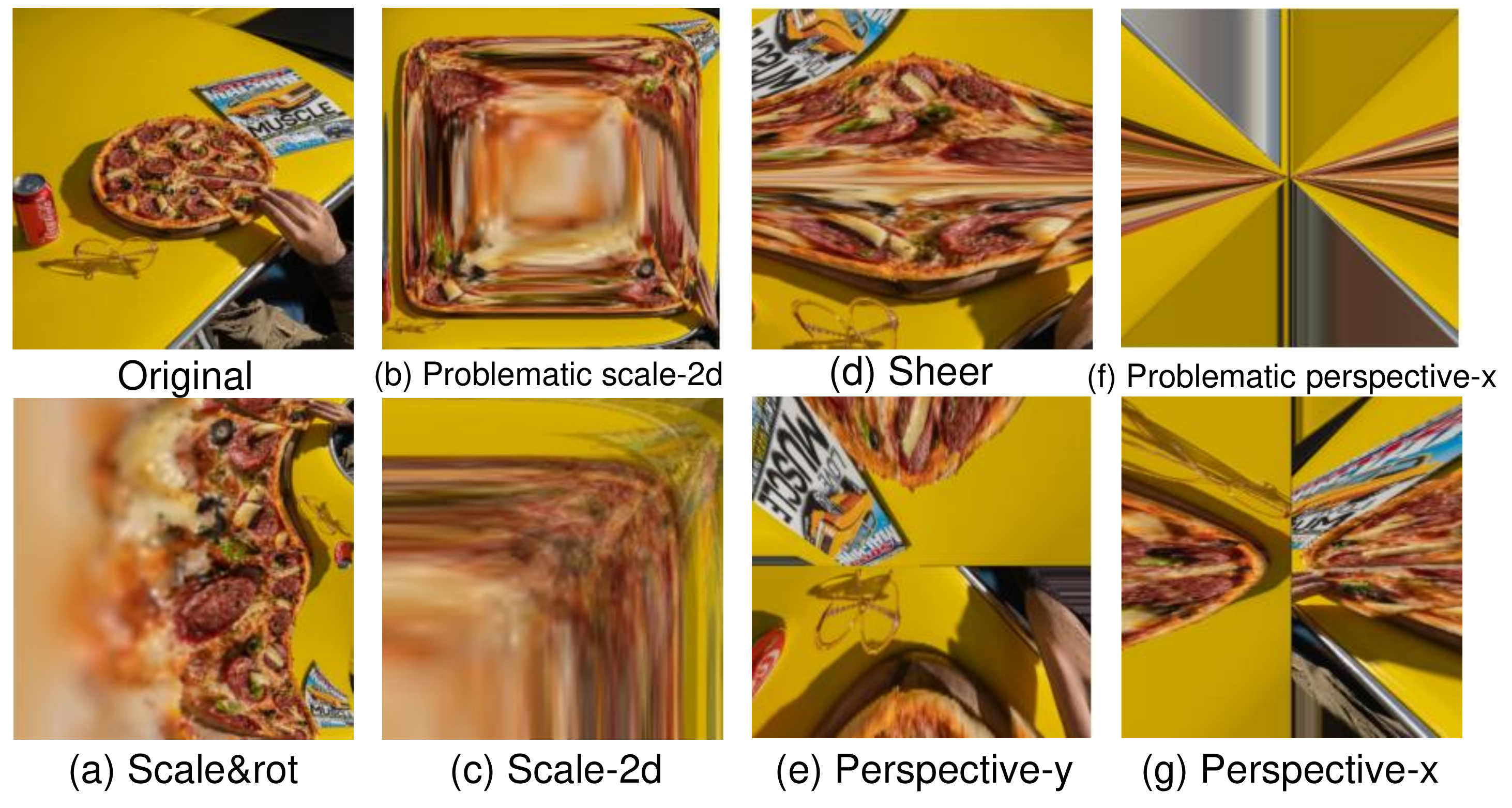}
    \vspace{-0.2in}
    \caption{ The examples after using the warping function for the original image.}
    \label{fig:warping_examples}
    \vspace{-0.2in}
\end{figure}

% \vspace{-2mm}
% \subsubsection{Scale and Rotation}
\vspace{-2mm}
\paragraph{Scale and Rotation}
Generally, CNNs are equivariant to the translation that is preserved after feature extraction. As a result, $w_0$ is an identical function and we omit it in our implementation.
% 	There is no need to design the warping function for translation. 
% 	various estimation methods depend on the tasks, and we introduce our examples in the experiment.  
For the scale and rotation groups, $\gamma$ represents the uniform scale, and $\theta$ denotes the in-plane rotation. As described in~\cite{Henriques2017WarpedCE, HDN}, the warping function $w_1$ for two Lie algebra coefficient parameters $b_3$ and $b_4$ is defined as:
%scale $\gamma$ and rotation $\theta$ is defined as:
\begin{small}
\begin{equation}
%	g_{\gamma,\theta}(\mathbf{\tau}) = 
\hspace{-0.5em}w_{1}(b_3,b_4) =\hspace{-0.2em} \mathbf{u}^{\prime T}\hspace{-0.2em}=\hspace{-0.2em}
\begin{bmatrix}
s^{\gamma^\prime} \|\mathbf{\tau}\|\cos(\arctan_2(\tau_2,\tau_1) + b_3) \\
s^{\gamma^\prime} \|\mathbf{\tau}\|\sin(\arctan_2(\tau_2,\tau_1) + b_3)
\end{bmatrix}.
\label{eq:warp_lp}
\end{equation}
\end{small}
where $s$ determines the degree of scaling. $\tau=(\tau_1,\tau_2)$ is the pivot, and $\gamma = s^{\gamma^\prime}$. The coordinates $\mathbf{u}^\prime = (u_1^\prime, u_2^\prime )$ denote the re-sampled point in image $I$, and $\arctan_2$ represents the standard 4-quadrant inverse tangent function. We set $\tau = (0,1)$ for convenience since $\tau$ can be any point except the origin. For the rest of the warp functions, $\tau$ can be used the same way, and we omit it from the warp function for simplicity. We have $\gamma = (s^{\gamma^\prime}=e^{\gamma^\prime \log s}) = e^{b_4} $. Let $s$ be a constant. Thus, estimating $\gamma^\prime$ is equivalent to finding the Lie algebra element $b_4$.
%\noindent\textbf{Example}
The range of the parameters should be consistent with the image size by scaling the coordinate in sampling. Therefore, we define the rescaled sampling function for the warped image $g_w^{n\times n}$ according to Eq.~\ref{eq:warp_lp} as $	( u_1^\prime,u_2^\prime)^T = [(\frac{n}{2})^{\frac{\mu_1}{n} } \cos(\frac{2\pi \mu_2}{n}  ), (\frac{n}{2})^{\frac{\mu_1}{n} } \sin(\frac{2\pi \mu_2}{n}  )]^T$.
%	\begin{equation}
%	\left\{
%	\begin{aligned}
%	&u = (\frac{n}{2})^{\frac{\mu_1}{n} } \cos(\frac{2\pi \mu_2}{n}  ) \\
%	&v = (\frac{n}{2})^{\frac{\mu_1}{n} } \sin(\frac{2\pi \mu_2}{n}  )
%	\end{aligned}
%	\right.
%	\label{eq:grid_lp}
%	\end{equation}
% \begin{equation}
% % 	\vspace{-0.05in}
% %	\left\{
% %	\begin{aligned}
% ( u_1^\prime,u_2^\prime)^T = [(\frac{n}{2})^{\frac{\mu_1}{n} } \cos(\frac{2\pi \mu_2}{n}  ), (\frac{n}{2})^{\frac{\mu_1}{n} } \sin(\frac{2\pi \mu_2}{n}  )]^T
% %	&u =  \\
% %	&v = 
% %	\end{aligned}
% %	\right.
% \label{eq:grid_lp}
% \end{equation}
% 	In Eq~\ref{eq:grid_lp},$\mathbf{\mu} =(\mu_1, \mu_2)$ is the sampled points in warped image. To make it clear,
Given the warped image with the size of $n\times n$, the warped range is limited by a circle whose radius is $\frac{n}{2}$ in the original image. Let $\hat{\mathbf{b}} = [\hat{b}_1, \hat{b}_2,..., \hat{b}_8]$ be the prediction of $\mathbf{b}$, $\hat{b}_3$ and $\hat{b}_4$ are recovered by $(\hat{b}_3,\hat{b}_4)=(\frac{2\pi\hat{\mu}_2}{n}, {\frac{\hat{\mu}_1}{n}}\log(\frac{n}{2}))$, where $(\hat{\mu}_1, \hat{\mu}_2)$ is the prediction of $(\mu_1, \mu_2)$. Fig.~\ref{fig:warping_examples} (a) shows the example warping functions for the scale and rotation. 
The mapping function performs on the warped image 
%$\mathcal{W}(I, \hat{\mathbf{H}}_s^{-1}\cdot \hat{\mathbf{H}}_t^{-1})$ 
$\mathcal{W}(I, [\hat{b}_1,\hat{b}_2])$ 
according to the estimated parameters from $E_1$.

\vspace{-2mm}
% \subsubsection{Aspect Ratio}
% \vspace{-1mm}
\paragraph{Aspect Ratio}
For group $\mathbf{H}_s$ which represents aspect ratio changes, its corresponding element of Lie algebra is $b_5$. Since there is a redundant dimension, we employ the warping function with two vertical directions \comment{$k_x\in(0,+\infty)$ and $k_y\in(0,+\infty)$} in order to double-check the parameter $b_5$. 
\comment{through the supervision on $k_x\to k_1$ and $k_y\to\frac{1}{k_1}$.}
The corresponding warp function is defined as follows,
%	\begin{equation}
%	g_{k_1^{\prime},1/k_1^{\prime}}(\mathbf{\tau}) = 
%	\begin{bmatrix}
%	s^{k_1} \\
%	s^{\frac{1}{k_1}} \\
%	\end{bmatrix}
%	\label{eq:warp_dblog}
%	\end{equation}
% 	\begin{equation}
%	g_{k_1^{\prime},1/k_1^{\prime}}(\mathbf{\tau}) = 	
% 	w_{2}(k_x,k_y) = 
% 	\begin{bmatrix}
% 	s^{k_x} , s^{k_y}
% 	\end{bmatrix}^T
% 	\label{eq:warp_dblog}
% 	\end{equation}
\begin{equation}
% 	\vspace{-0.1in}
%	g_{k_1^{\prime},1/k_1^{\prime}}(\mathbf{\tau}) = 	
w_{2}(b_5,-b_5) = 
\begin{bmatrix}
s^{k_x^\prime} , s^{k_y^\prime}
\end{bmatrix}^T.
\label{eq:warp_dblog}
\end{equation}
%\textcolor{blue}{Why we (not use)/(use) the $\tau$. how to represent the function with $tau$}
%\noindent\textbf{Quadrant problem} 
where \comment{$(k_x, k_y) = (s^{k_x^\prime},s^{k_y^\prime})$. Similarly, }$k_1 = (s^{k_x^\prime}=\exp({k_x^\prime\log s})) = \exp({b_5})$ and $1/k_1 = (s^{-k_x^\prime}=\exp({-k_x^\prime\log s})) = \exp({-b_5})$. Estimating $k_x^\prime$ and $k_y^\prime$ is actually to find the $b_5$ of $\mathfrak{sl}(3)$. In the image space, the range of parameters should be consistent with the image size by scaling the coordinate for sampling. The rescaled sampling function for scale estimation in both directions from Eq.~\ref{eq:warp_dblog} can be derived as $\mathbf{u}^{\prime T} = [(\frac{n}{2})^{\frac{2\mu_1}{n}}, (\frac{n}{2})^{\frac{2\mu_2}{n}} ]^T$.  
% \begin{equation}
% [u_1^\prime,u_2^\prime]^T =
% [(\frac{n}{2})^{\frac{\mu_1}{n}} ,
% (\frac{n}{2})^{\frac{\mu_2}{n}} ]^T
% \label{eq:grid_dblog}
% \end{equation}
$b_5$ and $-b_5$ are recovered by $(\hat{b}_5, -\hat{b}_5)=(\frac{\hat{2\mu}_1}{n}\log(\frac{n}{2}), \frac{\hat{2\mu}_2}{n}\log(\frac{n}{2}))$.
Since the main task is usually related to an object, its center is treated as the origin of coordinates.
In Eq.~\ref{eq:warp_dblog}, $w_{2}\in(0,+\infty)^2$.
Thus, we overpass the other quadrants when $u_1<0$ or $u_2<0$. Fig.~\ref{fig:warping_examples} (b) shows the resulting image. To account for the general case, we flip the other quadrant image to the positive quadrant and upsample it to the original image size of $n \times n$. 
This changes the size of the warped image $g_w^{n\times n\times 4}$.
Fig.~\ref{fig:warping_examples} (c) shows the example result of $g_w$.
Under the proposed framework, the mapping function performs on the warped image 
%$\mathcal{W}(I, \hat{\mathbf{H}}_s^{-1}\cdot \hat{\mathbf{H}}_t^{-1})$ 
$\mathcal{W}(I, [\hat{b}_1,\hat{b}_2,\hat{b}_3,\hat{b}_4])$ 
according to the estimated parameters.

\vspace{-2mm}
% \subsubsection{Shear}
% \vspace{-1mm}
\paragraph{Shear}
Shear transformation, also known as shear mapping, displaces each point in a fixed direction. According to the following equation on point $\mathbf{u}$, it can be found that shear is caused by the translation of each row in the original image, in which the translation degree is uniformly increased along with the column value. $\mathbf{u}^*$ is the transformed points as below  
\begin{equation}
\mathbf{u}^* = \mathbf{H}_{sh} \cdot \mathbf{u}^T = 
\begin{bmatrix} 
u_1+k_2 u_2 ,& 
u_2
\end{bmatrix}^T.
\end{equation}
Inspired by the fact that the arc length of each concentric circle with the same angle increases by the radius uniformly, it is intuitive to project the lines onto a circle arc so that the sheer can be converted into rotation. Similar to the warping in Eq.~\ref{eq:warp_lp}, the rotation is eventually formulated into the translation estimation. The warping function for shear can be derived as follows:
\begin{equation}
%	g_{k_2,u_2}(\mathbf{\tau}) = 
w_{3}(b_6, b_\epsilon) = 
\begin{bmatrix}
b_6b_\epsilon, & 
b_\epsilon 
\end{bmatrix}^T.
\label{eq:warp_sh}
\end{equation}
% 	}
%\noindent\textbf{Quadrant problem} 
%xxx seems no problem but happens in the experiment 
%\textbf{Example}
where $b_\epsilon$ is the unchanged coordinate for one-parameter group. In the case of a real image, the rescaled sampling function for shearing from Eq.~\ref{eq:warp_sh} becomes 	$\mathbf{u}^{\prime T} = \begin{bmatrix}
\frac{2}{n}(\mu_1 *\mu_2),& \mu_2
\end{bmatrix}^T $. 
% \begin{equation}
% %	\setlength{\arrayrowsep}{1.2pt}
% [u_1^\prime,u_2^\prime]^T = \begin{bmatrix}
% \frac{2}{n}(\mu_1 *\mu_2),& \mu_2
% \end{bmatrix}^T  
% \label{eq:grid_sh}
% \end{equation}
The estimated $\hat{b}_6$ is recovered by $\hat{k}_2 = \hat{\mu}_1$.
Finally, the warping function performs on the warped image
% $\mathcal{W}(I, \hat{\mathbf{H}}_{sc}^{-1}\cdot \hat{\mathbf{H}}_s^{-1}\cdot \hat{\mathbf{H}}_t^{-1})$ 
%
$\mathcal{W}(I, [\hat{b}_1, \hat{b}_2, ..., \hat{b}_5])$  
according to the estimated parameters.
Fig.~\ref{fig:warping_examples} (d) shows an example of sheering in the horizontal direction.
%	, where $n=127$. 

%\begin{figure}[h]
%	\centering 
%	\includegraphics[scale=0.3]{fig/perspective-example.png}
%	\caption{ The perspective warping example.}
%	\label{fig:persp-warp}
%\end{figure}

\vspace{-2mm}
% \subsubsection{Perspective}
% \vspace{-1mm}
\paragraph{Perspective}
The two elements $\nu_1$ and $\nu_2$ reflect the perspective distortion of an image, which is not the same as the previous one due to view change. 
\begin{small}
\begin{equation}
\hspace{-0.5em}\mathbf{u}^* \hspace{-0.35em}=\hspace{-0.3em} \mathbf{H}_{p}  \mathbf{u}^T \hspace{-0.35em}=\hspace{-0.3em}  \mathbf{H}_{p2} \mathbf{H}_{p1} \mathbf{u}^T \hspace{-0.35em}=\hspace{-0.35em}
\begin{bmatrix} 
\frac{u_1}{\nu_1 u_1 + \nu_2 u_2+1} ,& \hspace{-0.7em}
\frac{u_2}{ \nu_1 u_1 + \nu_2 u_2+1}
\end{bmatrix}^T
\label{eq:perspective-action}
\end{equation}
\end{small}
% \hspace{-0.1em}\mathbf{A_8}\hspace{-0.45em} = \hspace{-0.45em}
where $\mathbf{H}_p$ denotes the transformation for perspective change. From the action of the group $\mathbf{H}_{p}$ in Eq.~\ref{eq:perspective-action}, it can be found 
% 	it can be found that {\bf ??? its property are the line $u_2=0$ and $u_1=0$ does not move.} Therefore, we name it as ``central perspective group". 
% 	Another property is 
that the slope of any point does not change after the transformation. 
% 	Due to the surjective of the variables $\mu$ and $x$. 
$b_7$ and $b_8$ are entangled in Eq.~\ref{eq:perspective-action}. To make it clear, we design two one-parameter warp functions to account for the perspective changes of two groups $\mathbf{H}_{p1}$ and $\mathbf{H}_{p2}$.
    \begin{equation}
    \hspace{-0.5em}w_{4}(b_7, b_\epsilon) = 
    \begin{bmatrix}
    \frac{1}{b_7}, &
    \frac{b_\epsilon}{b_7}
    \end{bmatrix}^T\hspace{-0.2em} ,\hspace{-0.1em}  
    w_{5}(b_\epsilon, b_8) = 
    \begin{bmatrix}
    \frac{b_\epsilon}{b_8}, &
    \frac{1}{b_8}
    \end{bmatrix}^T.
	\hspace{-0.2em}  
    \label{eq:warp_ps}
    \end{equation}
% }	

As the same output size is required in sampling, the warp function in Eq.~\ref{eq:warp_ps} for sampling can be derived as $\mathbf{u}^{\prime T}=\begin{bmatrix}
\frac{n}{2\mu_1}, &
\frac{\mu_2 n}{ 2\mu_1}
%	\nu_1^*\nu_1
\end{bmatrix}^T$ and $\mathbf{u}^{\prime T}=\begin{bmatrix}
\frac{\mu_1 n}{ 2\mu_2},&
\frac{n}{2\mu_2}
\end{bmatrix}^T$.
% \begin{equation}
% [u_1^\prime, u_2^\prime]^T = 
% \begin{bmatrix}
% \frac{n}{2\mu_1}, &
% \frac{\mu_2 n}{ 2\mu_1}
% %	\nu_1^*\nu_1
% \end{bmatrix}^T,  
% [u_1^\prime, u_2^\prime]^T = 
% \begin{bmatrix}
% \frac{\mu_1 n}{ 2\mu_2},&
% \frac{n}{2\mu_2}
% \end{bmatrix}^T.
% \label{eq:grid_ps}
% \end{equation}
$\hat{b}_7 $ and $\hat{b}_8$ are recovered by $(\hat{b}_7, \hat{b}_8) = (\hat{\mu}_1, \hat{\mu}_2)$.
As depicted in Fig.~\ref{fig:warping_examples} (f), there exist serious distortions
when this sampling function is used. The larger the radius is, the more sparse the sampling points are. To tackle this issue, we select the patch near the center of the warped image. For transformation $\mathbf{H}_{p1}$, $w_4=[\frac{\phi_2 n}{2(\mu_1+sgn(\mu_1)\phi_1)}, \frac{u_2 n}{2(\mu_1+sgn(\mu_1\phi_1))}]^T$, where $sgn$ is the signum function. $\phi_1$ and $\phi_2$ are scaling factors.
% The above function acts on the warped image 
$w_4$ acts on the warped image 
$\mathcal{W}(I, [\hat{b}_1, \hat{b}_2,..,\hat{b}_6])$ and $w_5$ acts on the 
$\mathcal{W}(I, [\hat{b}_1, \hat{b}_2,..,\hat{b}_7])$ according to the estimated parameters.
%$\mathcal{W}(I, \hat{\mathbf{H}}_{sh}\cdot \hat{\mathbf{H}}_{sc}^{-1}\cdot \hat{\mathbf{H}}_s^{-1}\cdot \hat{\mathbf{H}}_t^{-1})$ 
%according to the estimated parameters and $H_{sc}$. 
Fig.~\ref{fig:warping_examples} (e,g) shows examples of the perspective warped image in two directions. 
%	 where $n=127$. 

\begin{figure}[t]
    \centering 
    \includegraphics[scale=0.40]{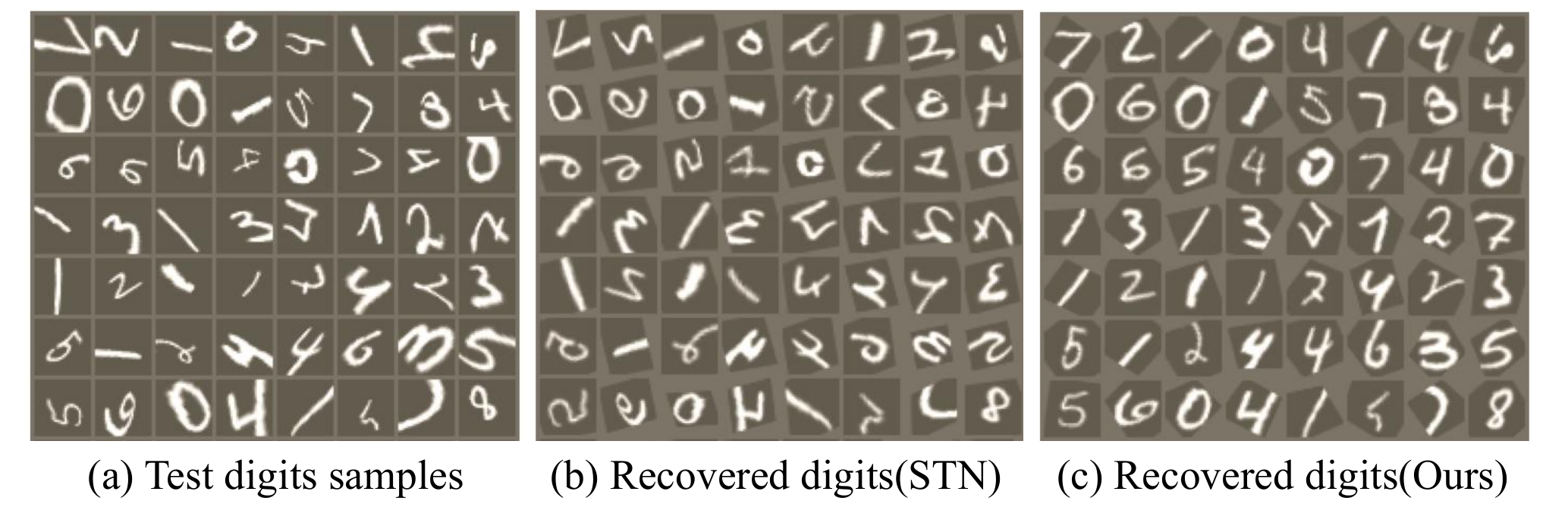}
    \vspace{-0.3in}
    \caption{ The visual results of the randomly selected MNIST testing after STN and WCN. }
    \label{fig:mnist-visualization}
    \vspace{-0.2in}
    % 	\vspace{-0.05in}
\end{figure}
% \vspace{-3mm}
\subsection{Implementation}\label{subsec:implementation}
% \vspace{-1mm}
The proposed WCN is designed to estimate the transformation with regard to the implicit or explicit reference image/object, which depends on the specific task. We give more details of structure in Section~\ref{sec:experiments} for two tasks\comment{~\ref{sec:implementation_details}}. The main idea is to recover the transformation parameters in the Lie subalgebras. To accomplish this goal, we add supervision to these elements through the robust loss function (i.e., smooth L1) defined in~\cite{FastRCNN}. As for the classification, we adopt the cross-entropy loss function. Another problem is that there are no training datasets with these parameter labels. We thereby augment the training datasets with the objects' bounding boxes and class labels. The augmentation scheme simulates the estimation process completely. The translation estimator depends on the requirements of the specific tasks. For those tasks without the explicit template, we directly use convolutional layers and linear layers to predict the translation. For the tasks having the template, we simply apply the cross-correlation to predict translation as is widely used in object tracking. In POT and S-COCO-Proj testing, we use the same translation estimator as HDN~\cite{HDN}.
More details can be found in Appendix B.
% We give our network details for MNIST-Proj testing in Appendix B.
% \textcolor{blue}{You might discuss about the choice of translation estimator?}
	
%------------------------------------------------------------------------
\section{Experiments}\label{sec:experiments}

To demonstrate the effectiveness of our proposed approach, we evaluate the WCN framework on three different learning tasks including classification, planar object tracking, and homography estimation.
All of them need to recover the underlying homography of the object. To this end, we have conducted the experiments on three datasets including MNIST-Proj, POT, and Synthetic COCO. 
% More results and analyses can be found in the appendix. %We modify the network structure to account for the different tasks.
\begin{figure*}[htbp]
    \centering 
    \includegraphics[scale=0.47]{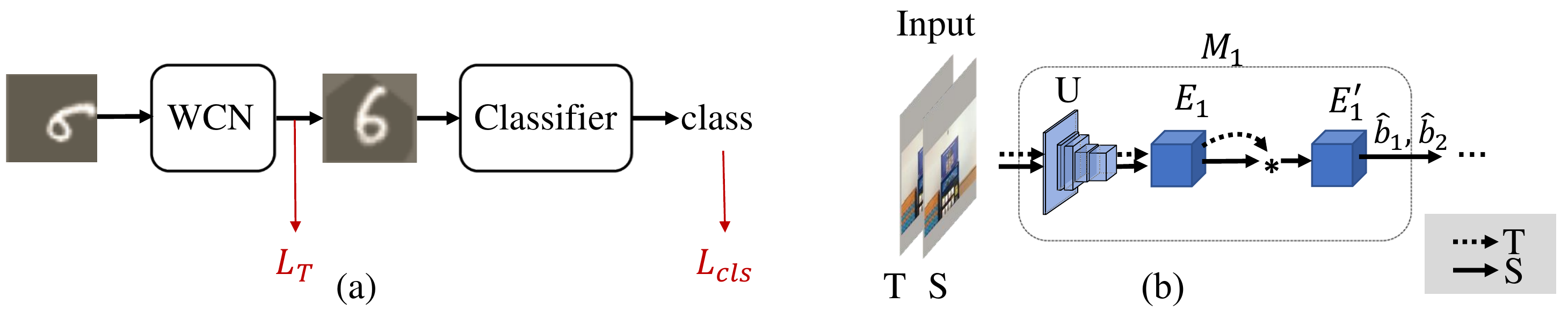}
    \vspace{-0.1in}
    \caption{The pipeline of our method for MNIST-Proj digits recognition and planar object tracking tasks. (a): MNIST-Proj digits recognition. $\mathcal{L}_T$: transformation loss, $\mathcal{L}_{cls}$: classification loss. (b): Planar object tracking. This figure only shows changes in a basic component $M_i$, $E_1$ and $E_1'$ are the convolution layers. Here, we use ResNet50 for backbone U. T is the template patch, and the dashed line denotes its data flow. Moreover, S is the search patch. and the solid line represents its data flow. }
    \label{fig:mnist-pipeline}
    \vspace{-0.2in}

\end{figure*}

% 	\begin{figure}[htbp]
% 	\centering 
% 	\includegraphics[scale=0.45]{fig/mnist-pipeline.png}
% 	\caption{The pipeline for mnist digits recognition.}
% 	\label{fig:mnist-pipeline}
% 	\end{figure}
% 	\begin{figure}[htbp]
% 		\centering 
% 		\includegraphics[scale=0.45]{fig/tracking-pipeline.png}
% 		\caption{ The pipeline for planar object tracking. $E_1$ and $E_1'$ are the convolution layers.}
% 		\label{fig:pot-pipeline}
% 	\end{figure}

% \vspace{-2mm}
\subsection{Classification on MNIST-Proj}
% \vspace{-1mm}

MNIST handwriting dataset~\cite{Mnist} is a small testbed for digits classification. We perform the experiment on it to show the effectiveness of WCN on the classification tasks. Specifically, we generate the MNIST-Proj dataset by augmenting the data in the training process with projective transformation. 
% The transformation parameters are randomly sampled from $\mathbf{x}$, where $\theta\in[1.5\mathrm{rad}, 1.5\mathrm{rad}]$, $\gamma\in[1/1.4, 1.4]$, $k_1\in[-0.3, 0.3]$, $k_2\in[-0.3, 0.3]$, $\nu_1\in[-0.2, 0.2]$, and $\nu_2\in[-0.2, 0.2]$. 
The testing dataset has 10,000 digits images and the size of samples is $28\times28$.

 \begin{table}[t]
\small
% \captionsetup{justification=centering,font=,font=footnotesize}
%\resizebox{0.3\textwidth}{!}{
% \begin{center}
% \begin{tiny}
    \centering
    \begin{tabular}{llc}
        \toprule
        %			\multicolumn{2}{c}{Part}                   \\
        \cmidrule(r){1-3}
        \bf Methods              & \bf Error ($\%$)  & \bf Time(ms) \\
        \midrule
        % 			Naive(\cite{LeNet})     & 11.48 ($\pm1.42$)     \\
        L-conv~\cite{LConv}   & 19.16 ($\pm$1.84)  &1.81 \\
        homConv~\cite{Macdonald2021EnablingEF}    & 14.72 ($\pm$0.72)  &105.7 \\
        PDO-econv~\cite{shen2020pdo} & 1.66 ($\pm$0.16) &\bf 0.14 \\
        LieConv~\cite{finzi2020generalizing})    & 2.7 ($\pm$0.74) &\textbackslash{}  \\
        PTN~\cite{PTN} & 2.45($\pm$0.66) & \textbackslash{}  \\
        STN~\cite{STN} & 0.79 ($\pm$0.07)  & 0.20\\
        \bf WCN~(Ours)                & \textbf{0.69} ($\pm$0.09)  & 0.42\\
        \bottomrule
    \end{tabular}
\vspace{-0.05in}
\caption{MNIST-Proj results.}

% \end{tiny}
%    \caption{Graph separation~\cite{balcilar2021breaking,ACGL-IJCAI21}.}
\label{tab:mnist-comp}
\vspace{-0.25in}
\end{table}

Usually, a homography recovery-based method requires a template reference.
For the classification problems, there is no explicit reference object to learn. Inspired by the congealing tasks~\cite{LearnedMiller2006DataDI},
% 	and easiness in learning from the Lie algebras as we described, 
we learn an implicit template pose, where the template is the upright digits in MNIST. As shown in Fig.~\ref{fig:mnist-pipeline}, the pipeline consists of two components. We first recover the transformation of the digit, and then employ the classifier to predict its class label. We add the supervision both on estimating the transformation parameters using loss function $\mathcal{L}_T$ and image class with loss function $\mathcal{L}_{cls}$. Thus, the total loss is $\mathcal{L}_T +\lambda \mathcal{L}_{cls}$, where $\lambda$ is the weight parameter to trade-off two terms.

The error rate is adopted as the metric for evaluation, which is calculated by the total wrongly predicted sample number divided by the total sample number. In Table~\ref{tab:mnist-comp}, our proposed framework outperforms six other methods. We use the official implementations for those methods,
and all the methods are trained with perspective transform augmentation. Although L-conv~\cite{LConv} is built based on Lie algebra, it performs inferior to the other methods when the group is fixed. 
% 	and where no change is added to their training settings and architectures 
% and We only change the dataset to MNIST-Proj. 
Especially, its results become worse when the center of the digits deviates from the image center. homConv~\cite{Macdonald2021EnablingEF} is theoretically equivariant to the $SL(3)$ group, nevertheless, it is not invariant at the feature level. This may lead to difficulty in identifying the digit class. PDO-econv~\cite{shen2020pdo} and PTN~\cite{PTN} handle the rotation well, yet we still attain a lower error rate.
Moreover, as shown in Fig.~\ref{fig:mnist-visualization}, 
% compared to its performance, estimate the transformation
the visual results show the advantage of our proposed WCN over STN in recovering the homography, which can be utilized directly in other tasks like homography estimation.
%	\begin{wrapfigure}[13]{r}{0.5\textwidth}
%		\vspace{-10pt}
%		\begin{center}
%			\includegraphics[width=0.5\textwidth]{figures/exp_plots/inv_and_m_f.pdf}
%		\end{center}
%		\vspace{-10pt}
%		\caption{Invariance error (left); $m_{_\gF}$-$m_{_G}$ (right).}\label{fig:inverr_m_f}
%	\end{wrapfigure}

%	\begin{figure*}[t]
%		\vspace{-0.25in}
%		\centering
%		\subfigure{
%			\begin{minipage}[t]{0.40\linewidth}
%				\centering
%				\includegraphics[width=1\linewidth]{fig/PD-precision.pdf}
%				%\caption{fig1}
%			\end{minipage}%
%		}%
%		\subfigure{
%			\begin{minipage}[t]{0.40\linewidth}
%				\centering
%				\includegraphics[width=1\linewidth]{fig/PD-homography.pdf}
%				% 			\includegraphics[width=1\linewidth]{figs/experiment/All-seq-precision.pdf}
%				%				\caption{fig2}
%			\end{minipage}
%		}%
%		\centering
%		\vspace{-0.25in}
%		\caption{Comparisons on POT. The left subfigure plots the Precision under different thresholds based on corner offsets, and the left subfigure plots the Success rate based on homography discrepancy.}
%		\label{fig:pot_compare}
%		\vspace{-0.2in}
%	\end{figure*}
	
% \vspace{-2mm}
\subsection{Planar Object Tracking on POT}
% \vspace{-2mm}
% The objective of planar object tracking is to estimate the homography from the template to the search image. 
POT~\cite{POT} is a mainstreamed planar object tracking dataset that contains 210 videos of 30 planar objects sampled in the natural environment.
We select the videos officially divided with perspective changes as our testing datasets to evaluate the performance in perspective transformation estimation. The state-of-the-art visual object trackers and planar object tracking approaches~\cite{HDN, Ocean} show that it is easy to predict the offset of an object through the cross-correlation\cite{SiamFC}. Thus, we employ it as the parameter estimator in our framework. As shown in Fig~\ref{fig:mnist-pipeline}, there are two inputs for WCN, where the correlation is used to estimate the Lie algebra parameters. The only difference is the perspective $b_7$ and $b_8$ estimation, since the offset value in the warped image is much smaller than other parameters. We hence directly estimate the pseudo-translation, which shares the higher accuracy.

% \begin{figure*}
%   \centering
%   \begin{subfigure}{0.68\linewidth}
%     \fbox{\rule{0pt}{2in} \rule{.9\linewidth}{0pt}}
%     \caption{An example of a subfigure.}
%     \label{fig:short-a}
%   \end{subfigure}
%   \hfill
%   \begin{subfigure}{0.28\linewidth}
%     \fbox{\rule{0pt}{2in} \rule{.9\linewidth}{0pt}}
%     \caption{Another example of a subfigure.}
%     \label{fig:short-b}
%   \end{subfigure}
%   \caption{Example of a short caption, which should be centered.}
%   \label{fig:short}
% \end{figure*}

% \begin{wrapfigure}[11]{r}{0.7\textwidth}
\begin{figure}
    %	\begin{figure*}[t]
    % \vspace{-0.35in}
    \centering
     \begin{subfigure}{0.47\linewidth}
        % \begin{minipage}[t]{0.50\linewidth}
            \centering
            \includegraphics[width=1\linewidth]{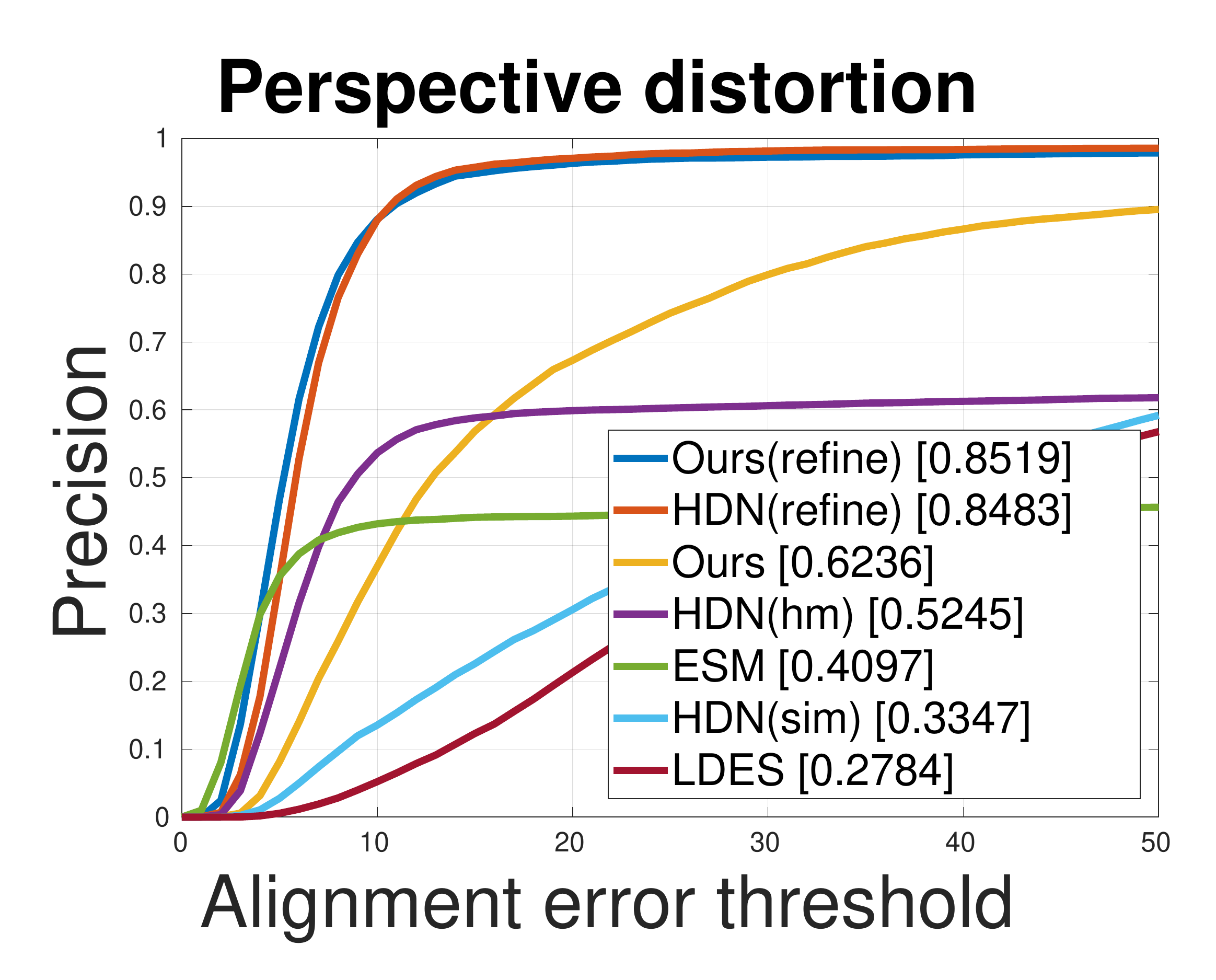}
            %\caption{fig1}
        % \end{minipage}%
  \end{subfigure}
     \begin{subfigure}{0.45\linewidth}
        % \begin{minipage}[t]{0.48\linewidth}
            \centering
            \includegraphics[width=1\linewidth]{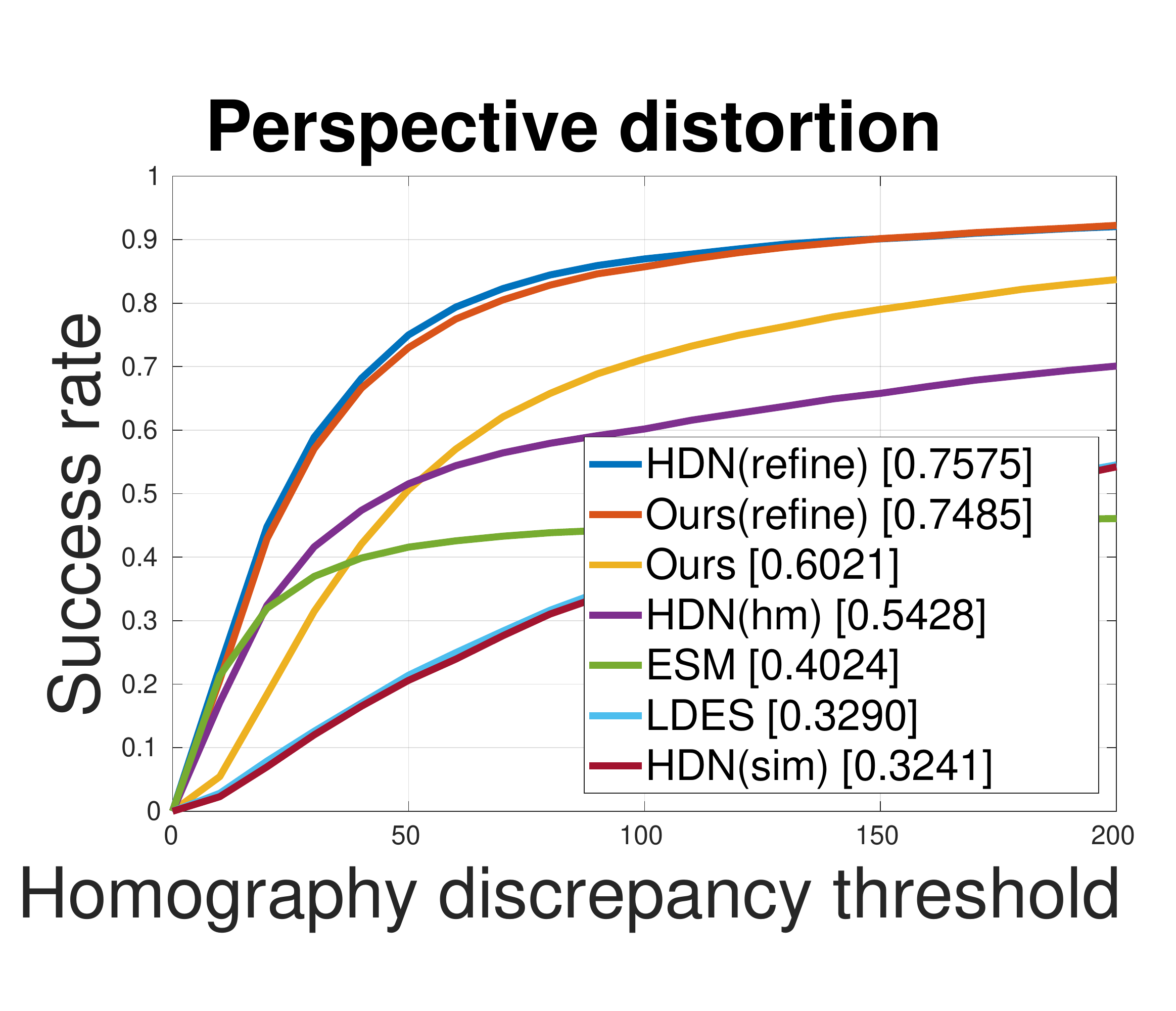}
            % 			\includegraphics[width=1\linewidth]{figs/experiment/All-seq-precision.pdf}
            %				\caption{fig2}
        % \end{minipage}
  \end{subfigure}
    \centering
    \vspace{-0.15in}
%		\caption{Comparisons on POT. The left subfigure plots the Precision under different thresholds based on corner offsets, and the right subfigure plots the Success rate based on homography discrepancy.}
    \caption{Comparisons on POT. Precision~(left), Success rate~(right).}
    \label{fig:pot_compare}
    \vspace{-0.2in}
    %	\end{figure*}
% \end{wrapfigure}
\end{figure}

% \begin{wrapfigure}[11]{r}{0.7\textwidth}
%     %	\begin{figure*}[t]
%     \vspace{-0.35in}
%     \centering
%     \subfigure{
%         \begin{minipage}[t]{0.50\linewidth}
%             \centering
%             \includegraphics[width=1\linewidth]{fig/PD-precision.pdf}
%             %\caption{fig1}
%         \end{minipage}%
%     }%
%     \subfigure{
%         \begin{minipage}[t]{0.48\linewidth}
%             \centering
%             \includegraphics[width=1\linewidth]{fig/PD-homography.pdf}
%             % 			\includegraphics[width=1\linewidth]{figs/experiment/All-seq-precision.pdf}
%             %				\caption{fig2}
%         \end{minipage}
%     }%
%     \centering
%     \vspace{-0.25in}
% %		\caption{Comparisons on POT. The left subfigure plots the Precision under different thresholds based on corner offsets, and the right subfigure plots the Success rate based on homography discrepancy.}
%     \caption{Comparisons on POT. Precision~(left), Success rate~(right).}
%     \label{fig:pot_compare}
%     % \vspace{-0.1in}
%     %	\end{figure*}
% \end{wrapfigure}
We choose three methods that directly estimate the Lie algebra coefficient elements for comparison. LDES~\cite{LDES} takes advantage of the log-polar coordinate to estimate the similarity transform. HDN~\cite{HDN} employs a deep network to estimate the similarity parameters, in which a corner offsets estimator is used to refine the corner. Thus, we divide HDN into two parts to compare the estimation performance fairly. ESM directly estimates 8 Lie algebra elements in images whose generators are different from ours. They compose these elements in one group, which does not share any equivariance in their structure. 

As defined in \cite{POT}, we adopt two metrics for evaluation, including precision and homography success rate. Precision is defined as the percentage of frames whose alignment error is smaller than the given threshold. The alignment error is calculated as the average of four points $L_2$ distance between the predicted polygon and the
ground truth label. Success rate
describes the percentage of frames whose homography discrepancy score is less than a threshold. The leaderboard shows the average precision and success rate, which denote the average precision and success rate of all error thresholds. 

The experimental result is shown in Fig.~\ref{fig:pot_compare}. Compared with other trackers, our proposed method achieves a higher average precision and success rate. It has a much higher precision when the error is larger~(up to 90\%), which indicates the high robustness of our method with 2D perspective transformation. This supports the homography learning in the proposed method. Our estimation is a bit coarse as the proposed framework is not specially designed for the tracking task. 
Meanwhile, with the refinement component~(HDN(hm)), our WCN achieves the comparable performance for two metrics and higher precision when the error threshold is low.
This is because the minor residual estimation error can be compensated easily when the large transformation is estimated correctly.
More experiments and analyses are provided in Appendix C.\comment{~\ref{sec:appendix_experiments}.}
\begin{table}[t]
\scriptsize
    \centering
    % \captionsetup{justification=centering,font=,font=footnotesize}
    %\resizebox{0.3\textwidth}{!}{
        % \begin{center}
        % \begin{tiny}
            \begin{tabular}{lcc}
                \toprule
                %			\multicolumn{2}{c}{Part}                   \\
                %		\cmidrule(r){1-2}
                %		Methods              & MACE\\		
                \cmidrule(r){1-3}
                %					\multicolumn{3}{l}{10k}
                \bf Methods              & \multicolumn{2}{c}{\bf MACE}  \\
                & \bf Middle Aug. & \bf Large Aug. \\
                \midrule
                Content-Aware~\cite{Zhang2020ContentAwareUD} &40.57 & 56.57 \\
                HomographyNet~\cite{DeTone2016DeepIH} &19.17 & 35.59 \\
                PFNet~\cite{Zeng2018RethinkingPH}   & 11.86  & 25.30\\
                PFNet+biHomeE~\cite{Koguciuk2021PerceptualLF} & 12.62 &33.12 \\
                \bf Ours                & \textbf{10.23}  & \textbf{17.73} \\
                \hline
                PFNet (wo.mask )    & 2.45  & 13.84 \\
                Ours (wo.mask )     & 6.29  & 11.31 \\
                \bf Ours+PFNet (wo.mask )  & \textbf{0.69}  & \textbf{2.35}\\
                \bottomrule
            \end{tabular}
            \vspace{-0.05in}
            \caption{S-COCO-Proj comparison.}
        % \end{tiny}
    %    \caption{Graph separation~\cite{balcilar2021breaking,ACGL-IJCAI21}.}
    \label{tab:scoco-comp}
    \vspace{-0.2in}
\end{table}	
\vspace{-2mm}
\subsection{Homography Estimation on S-COCO-Proj}
\vspace{-1mm}
S-COCO-Proj is a large synthetic homography estimation dataset based on COCO14~\cite{COCO}. We augment the COCO14 with middle and large transform augmentation and mask the corners to test homography estimation performance with the middle and large transformation and the occlusion influence~(see Appendix A\comment{~\ref{sec:appendix_warp_func}} for more details). 
This is because our method is robust rather than accurate due to several reasons like sampling density and the influence of different parameters as explained in  Appendix D. Besides, S-COCO~\cite{DeTone2016DeepIH} augments the data by moving the corners of the images, which mainly brings the perspective distortion and lacks other transformations like scale and rotation. 

Table~\ref{tab:scoco-comp} exhibits the performance when using S-COCO-Proj. We compare our WCN with the recent methods PFNet~\cite{Zeng2018RethinkingPH} in S-COCO-Proj. We adopt the standard MACE~(Mean Average Corner Error) metric to evaluate the performance. Even with the tracking procedure and inferring only once, our proposed approach outperforms the other four methods especially with large transformations.
% PFNet~\cite{Zeng2018RethinkingPH}, Content-Aware~\cite{Zhang2020ContentAwareUD}, HomographyNet~\cite{DeTone2016DeepIH} AND biHomeE~\cite{Koguciuk2021PerceptualLF}, especially with large transformations. 
Content-Aware~\cite{Zhang2020ContentAwareUD} is an unsupervised method, therefore performs worse with drastic transformations. biHomeE~\cite{Koguciuk2021PerceptualLF} adopts the perceptual loss for unsupervised learning. It is still limited by the unchanged predicted parameters when using the previous estimation genre. PFNet~\cite{Zeng2018RethinkingPH} and HomographyNet~\cite{DeTone2016DeepIH} use the offsets of the local points to recover the homography. Therefore, their results are inferior to our WCN when there are occlusions on the corners.
Moreover, we demonstrate the high accuracy performance of our proposed WCN as a robust homography representation for the SOTA method.
The only difference of the settings is that we remove the mask occlusion (wo.mask) for the testing to show the up-boundary accuracy performance.
%   to stress the lacking high accuracy problem of our WCN can be easily resolved, we add one more experiment without the mask on the corners of images. with the other method PFNet as the refinement module,
It can be observed that the performance of Ours+PFNet is significantly better and more accurate for middle and large transformations than using them separately. 
\vspace{-2mm}
\subsection{Robustness for Transformation}
\begin{figure}[t]
    \centering 
    \includegraphics[scale=0.38]{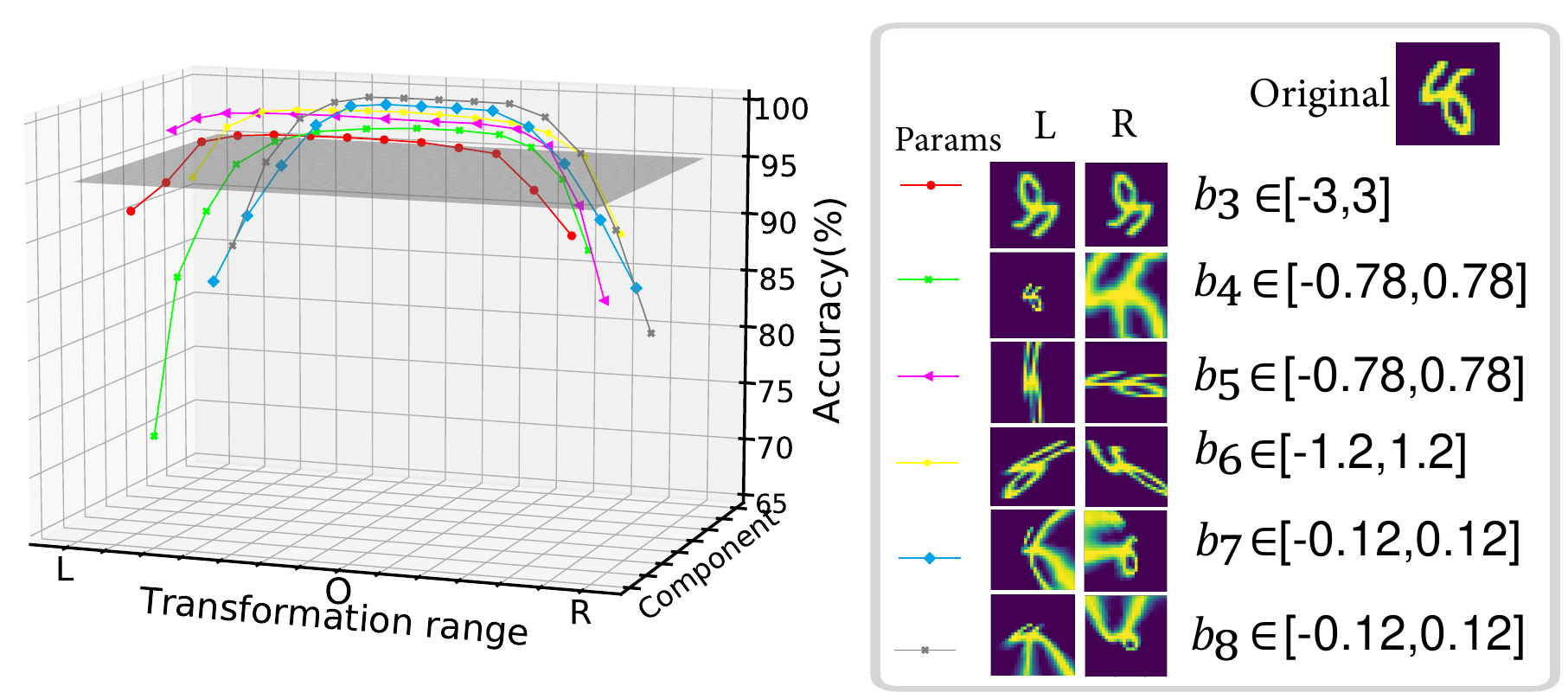}
    \vspace{-0.1in}
    \caption{ Range robustness for each parameter on MNIST-PROJ. }
    \label{fig:robust-range}
    % \vspace{-0.2in
    \vspace{-0.2in}
\end{figure}

\vspace{-1mm}
To evaluate the robustness of the proposed method under different transformation ranges, we further test with a wide range of parameters $\mathbf{b}$. As there are two directions for each $i_{th}$ parameter $b_i$, it is hard to analyze them together. We thereby conduct the experiment on each parameter separately on MNIST-Proj. Fig.~\ref{fig:robust-range} shows the result, where L is the left boundary for the transformation parameter, and R denotes the right boundary. We plot examples for every parameter resulting in transformed images. The gray surface marked the standard 95$\%$ accuracy level, our WCN achieves a large proportion over this threshold.  This confirms a satisfying upper bound for a large transformation range.

%%%%%%_-------two table side by side _------------%%%%%%%
\begin{figure}[t]
\vspace{0.1in}
\begin{minipage}[l]{0.26\textwidth}
    \small
    % \centering
    \begingroup
    \setlength{\tabcolsep}{2pt}
    \renewcommand{\arraystretch}{1} % Default value: 1
    % 		\vspace{-0.05in}
%		\caption{Ablation of backbone on MNIST-Proj. All the models except the Naive (wo.aug) are trained with augmentation}
%		\label{tab:mnist-proj}
    \scriptsize
    % \centering
    \begin{tabular}{lll}
        \toprule
        %			\multicolumn{2}{c}{Part}                   \\
        \cmidrule(r){1-3}
        \bf Methods           & \bf Network   & \bf Error ($\%$) \\
        \midrule
        Naive & LeNet5$^*$    & 11.48 ($\pm1.42$)     \\
        % Naive(wo.aug)     & LeNet5$^*$    & 59.37 ($\pm0.99$)     \\
        Navie             & ResNet18  & 0.87 ($\pm$0.13)  \\			
        STN  & LeNet5$^*$ & 4.00 ($\pm$0.35)\\
        STN               & ResNet18  & 0.79($\pm$0.07)  \\
        Ours              & LeNet5$^*$  & 3.05 ($\pm$0.33)  \\
        Ours              & ResNet18  & 0.69 ($\pm$0.09)  \\
        \bottomrule
    \end{tabular}
    \captionof{table}{Ablation of backbone \\on MNIST-Proj.}
    \label{tab:mnist-proj}
 \endgroup
\end{minipage}
\begin{minipage}[r]{0.18\textwidth}
    \scriptsize
    % \centering
    \begingroup
    \setlength{\tabcolsep}{2pt}
    \renewcommand{\arraystretch}{1} % Default value: 1
        \begin{tabular}{ll}
        \toprule
        %			\multicolumn{2}{c}{Part}                   \\
        \cmidrule(r){1-2}
        \bf Params              & \bf Precision\\
        \midrule
        $[b_1,b_2,...,b_8]$    & 62.4$\%$     \\
        $[b_1,b_2,...,b_6]$     & 49.5$\%$\\			
        $[b_1,b_2,...,b_5]$  & 39.8$\%$\\
        $[b_1,b_2,b_3,b_4]$       &  33.1$\%$\\
        $[b_1,b_2,b_3]$               & 18.8$\%$  \\
        $[b_1,b_2]   $            & 13.6$\%$   \\
        \bottomrule
    \end{tabular}
    \captionof{table}{Ablation on warp functions}
    \label{tab:ab-warp}
     \endgroup
\end{minipage}
\vspace{-0.2in}
\end{figure}
%%%%%%_-------two table side by side _------------%%%%%%%

% \vspace{-4mm}	
\subsection{Ablation Study}
% \vspace{-2mm}	
% There are few studies on learning the planar transformation equivariance or invariance for CNNs. Most of their implementations are not publicly available. 

For fair evaluation, we compare our proposed approach with four baseline methods and compare them with the same backbone. As shown in Table~\ref{tab:mnist-proj}, we use the mean error in the last five epochs to measure the performance. 
% All the methods are trained from scratch with the same number of epochs and learning rate. 
% A naive modified LeNet~\cite{LeNet} with the augmented data can obtain an acceptable result for MNIST-Proj. If there is no augmentation~(wo.aug), the accuracy drops drastically. This reveals that CNNs are data-driven. 
% It does not generalize well without the appearance of the patterns. 
When equipped with deeper convolution layers (ResNet18), the CNNs are able to classify the digits well even with large transformations. To fairly compare with STN~\cite{STN}, we use the same backbone for classification and achieve a lower error rate. With a five-layer CNN, our proposed approach outperforms STN by 1\%. There is no significant gap between WCN and STN using a deeper CNN backbone. This is because there is little space for those hard cases. As shown in Fig.~\ref{fig:mnist-visualization}, our visual results of the recovered transformations are much better than STN's. %This may reveal that the same pose of an object is not so important for classification. 

To evaluate the contribution from each warp function and the adaptability for different groups, we conduct the experiment on POT with different warped functions for transformation parameters. The results are shown in Table~\ref{tab:ab-warp}. It can be seen that the results are better with more warped functions and parameters. Besides, we can combine different warped functions freely.

\vspace{-2mm}
\section{Conclusion}
\vspace{-2mm}
In this paper, we proposed Warped Convolution Networks~(WCN) to effectively learn the homography by $SL(3)$ group and $\mathfrak{sl}(3)$ algebra with group convolution. 
Based on the warped convolution, our proposed WCN extended the capability of handling noncommutative groups and achieved to some extent equivariance.
To this end, six commutative subgroups within the $SL(3)$ group along with their warping functions were composed to form a homography. 
By warping the corresponding space and coordinates, the group convolution was accomplished in a very efficient way.
Extensive experiments showed that our proposed method is effective for representation learning.
The proposed approach achieved the highest performance for both estimation tasks and classification problem compared to the direct methods on POT and S-COCO-Proj, and STN on MNIST-Proj.

% ------------------------------------------------------
% Appendix
\clearpage
\appendix
{\noindent\Large\textbf{Appendix}}
\newline

	In this appendix, we first discuss the details of the warp function and analyze the influence of each parameter on the warped image.
	Then, the implementation details of our proposed method are provided, and more experimental details are introduced with additional results.
	Finally, we provide the proof of warping function property.
	
	\section{Warp Functions}\label{sec:appendix_warp_func}
	\label{sec:warp-func}
	For the warped convolution~\cite{Henriques2017WarpedCE}, the most ideal situation is that the group has commutative property with only two parameters. In this case, all the parameters are independent and the group convolution can be implemented as a warped function. However, it is impossible for both the affine group and projective group to have the same properties since they are not Abelian groups with more parameters. A transformation matrix can be represented as follows
	\begin{equation}
	\mathbf{H} = \exp(\mathbf{A}(\mathbf{b})) = \exp(\sum_{i=1}^{8}b_i\mathbf{A}_i),
	\end{equation}
	where $\mathbf{A}_i$ is the generator of the Lie algebra $\mathbf{A}$. $b_i$ is an element of the generator coefficients vector $\mathbf{b}$ in the real plane.
	For affine and projective group, it does not hold that $e^{\mathbf{A(b)}}e^{\mathbf{A(a)}} = e^{\mathbf{A}(\mathbf{a+b})}$. 
	Intuitively, this means the element of the Lie algebra loses its meaning in the image transform, while the corresponding one-parameter subalgebra still holds the property, e.g. rotation and scale.
	Besides, it does not satisfy the condition for Eq.~(6) in the main paper. Therefore, no warping function can be found for both affine and projective groups directly.
	
	It is worthy of discussing why not map the projective transformation onto 3D space to estimate the 6 independent parameters. 
	% 	There are three parameters of the rotation group SO(3). 
	The reason is that one cannot project the image into a particular camera view without the depth and camera intrinsic. Therefore, there is no way to warp the image like the log-polar coordinates for in-plane rotation~\cite{PTN}.
	In section 3.1 of the main paper, we follow the warped convolution~\cite{Henriques2017WarpedCE} and decompose the homography into 6 subgroups that can be predicted independently by pseudo-translation estimation according to the equivariance.
	% 	found that the networks are able to localize the object’s position very well for most of the vision tasks. 
	% 	Especially, with the transnational equivariance of the convolution, CNN demonstrates the capability of estimating translational parameters~\cite{TrackingNet, HDN} of the different objects.
	% 	This makes it possible to estimate the parameters as the translation.
	
\begin{figure*}[htbp]
		\centering
			\begin{subfigure}{0.24\linewidth}
				\centering
				\includegraphics[width=1\linewidth]{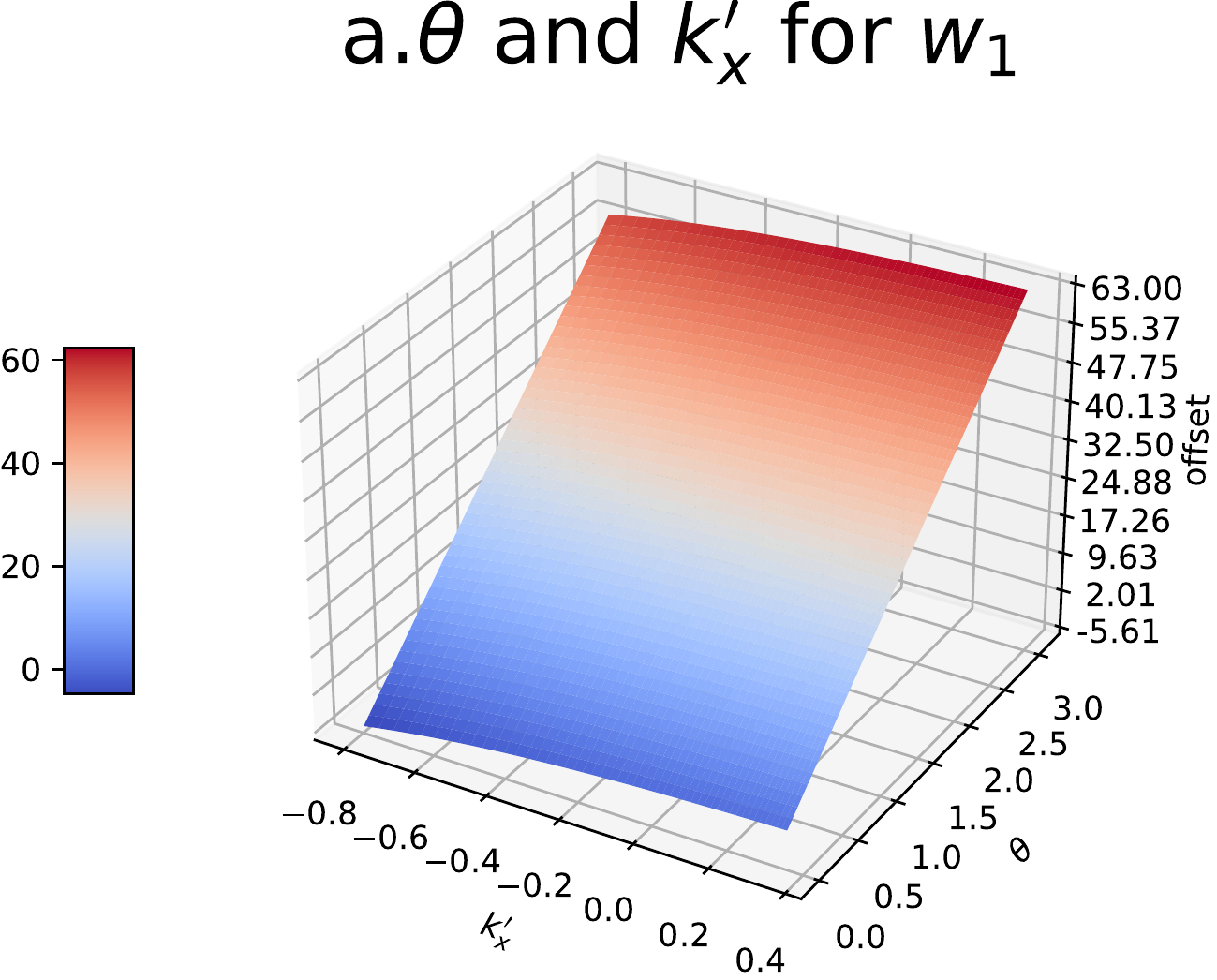}				%\caption{fig1}
			\end{subfigure}%
			\begin{subfigure}{0.24\linewidth}
				\centering
				\includegraphics[width=1\linewidth]{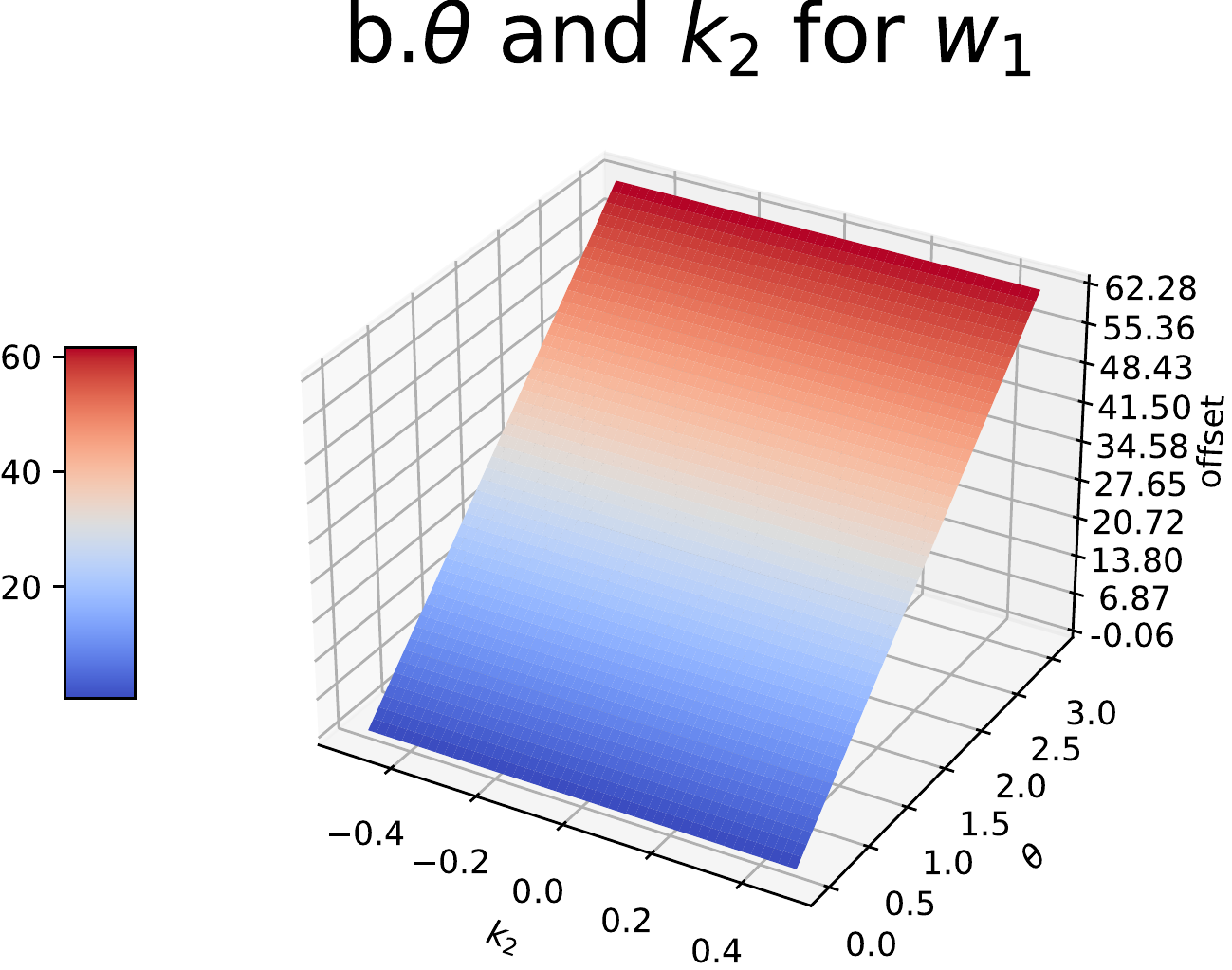}
				% 			\includegraphics[width=1\linewidth]{figs/experiment/All-seq-precision.pdf}
				%				\caption{fig2}
			\end{subfigure}
		\centering
			\begin{subfigure}{0.24\linewidth}
				\centering
				\includegraphics[width=1\linewidth]{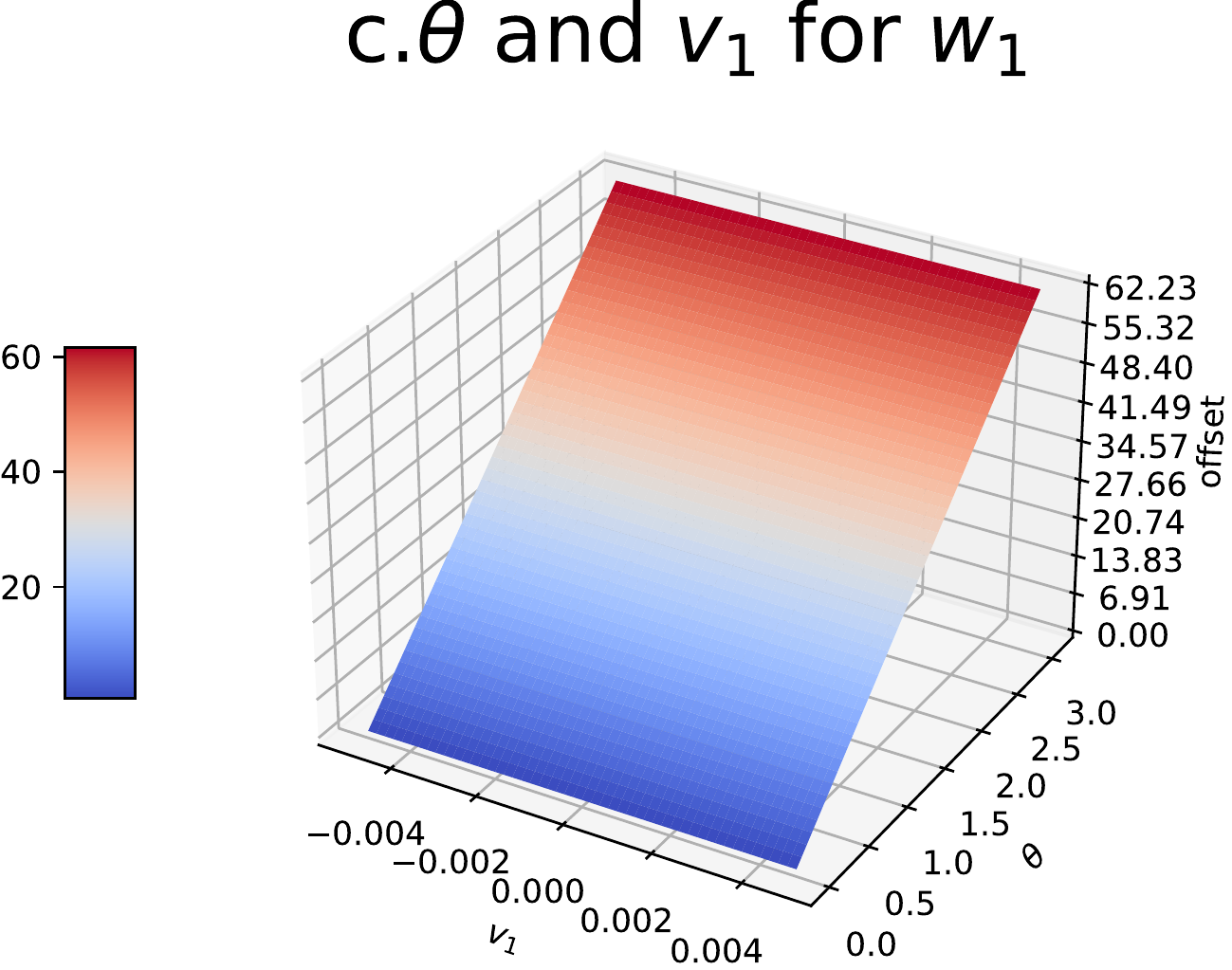}
				%\caption{fig2}
			\end{subfigure}
		\centering
			\begin{subfigure}{0.24\linewidth}
				\centering
				\includegraphics[width=1\linewidth]{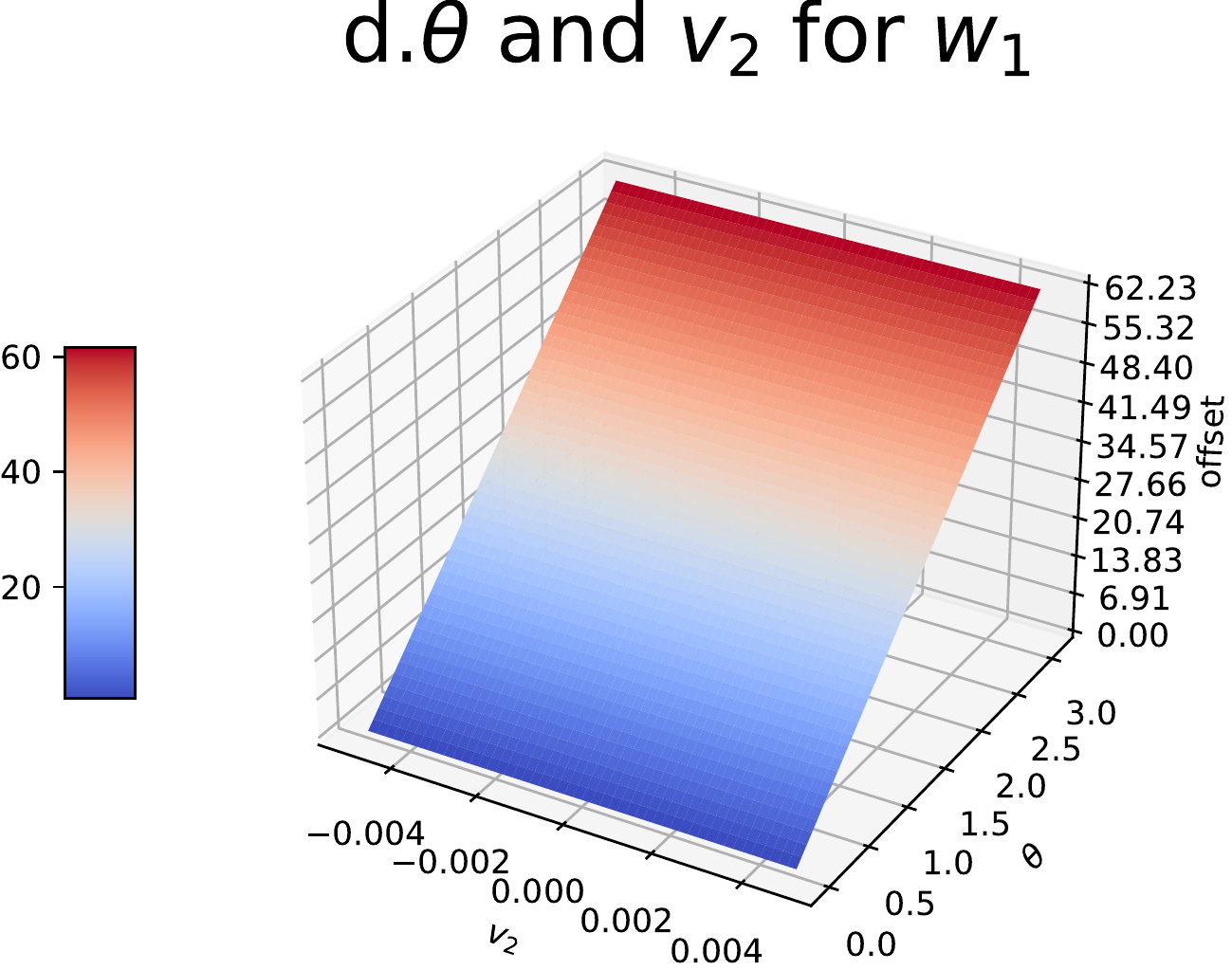}
				%\caption{fig2}
			\end{subfigure}
		
		\centering
			\begin{subfigure}{0.24\linewidth}
				\centering
				\includegraphics[width=1\linewidth]{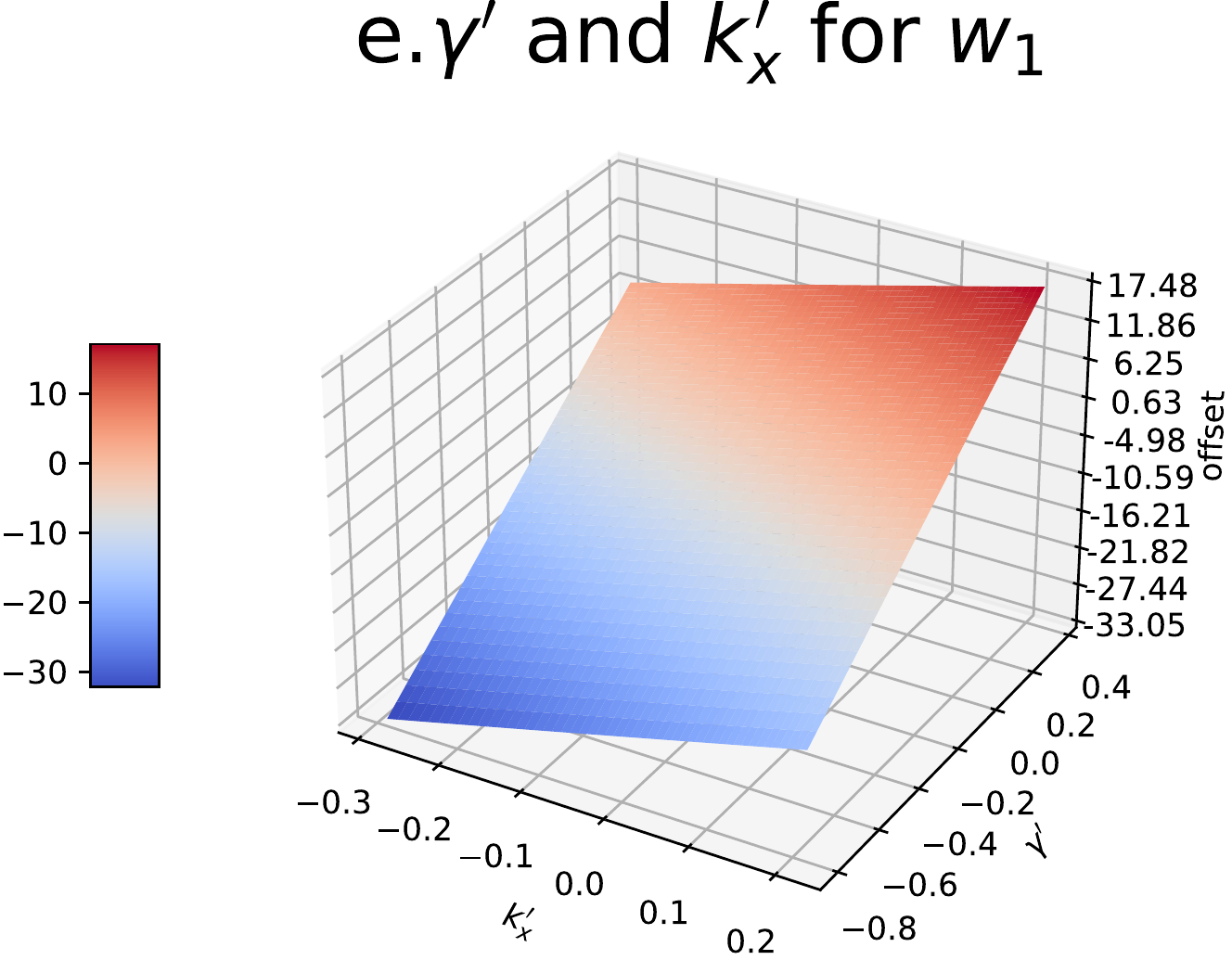}
				%\caption{fig2}
			\end{subfigure}
		\centering
			\begin{subfigure}{0.24\linewidth}
				\centering
				\includegraphics[width=1\linewidth]{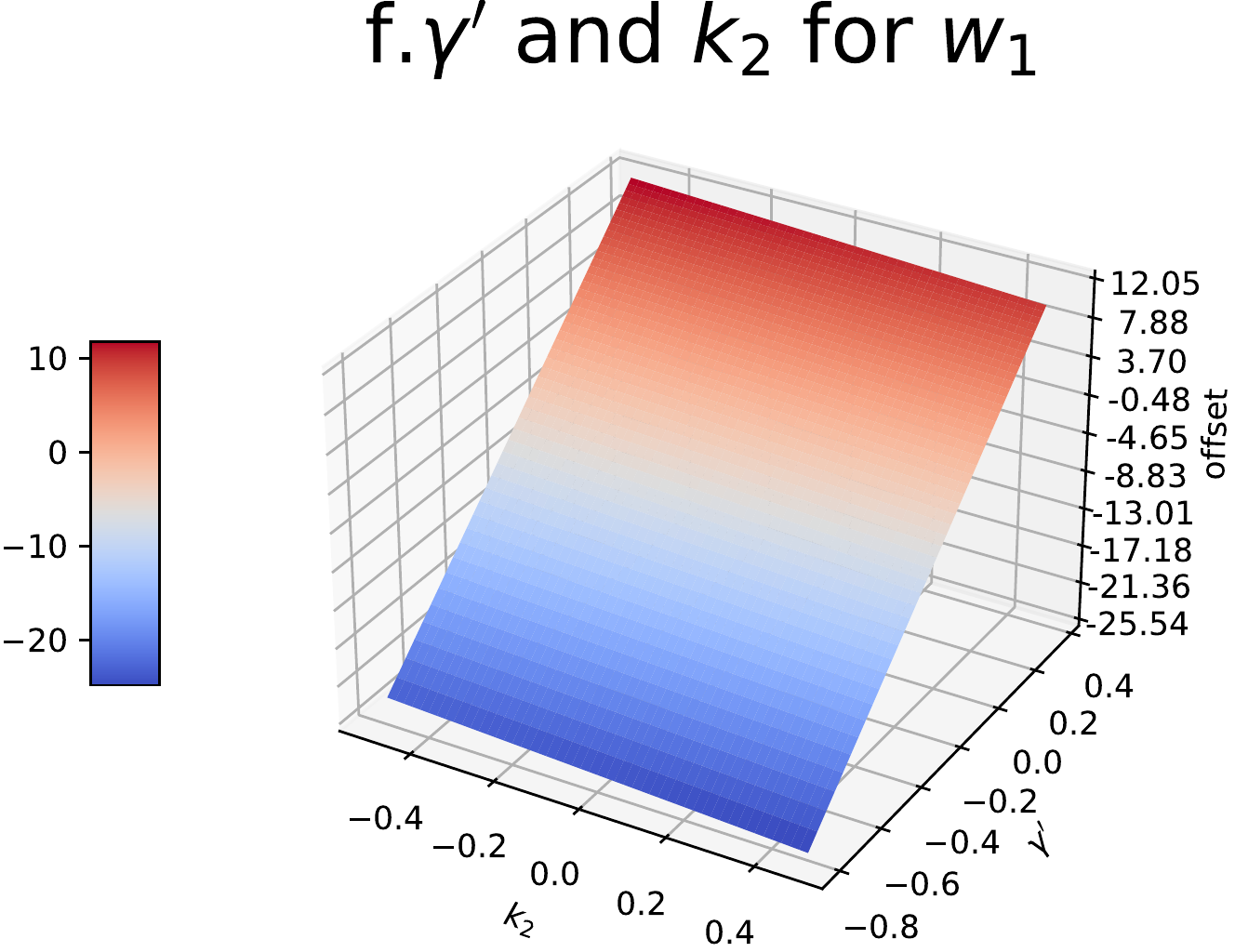}
				%\caption{fig2}
			\end{subfigure}
		\centering
			\begin{subfigure}{0.24\linewidth}
				\centering
				\includegraphics[width=1\linewidth]{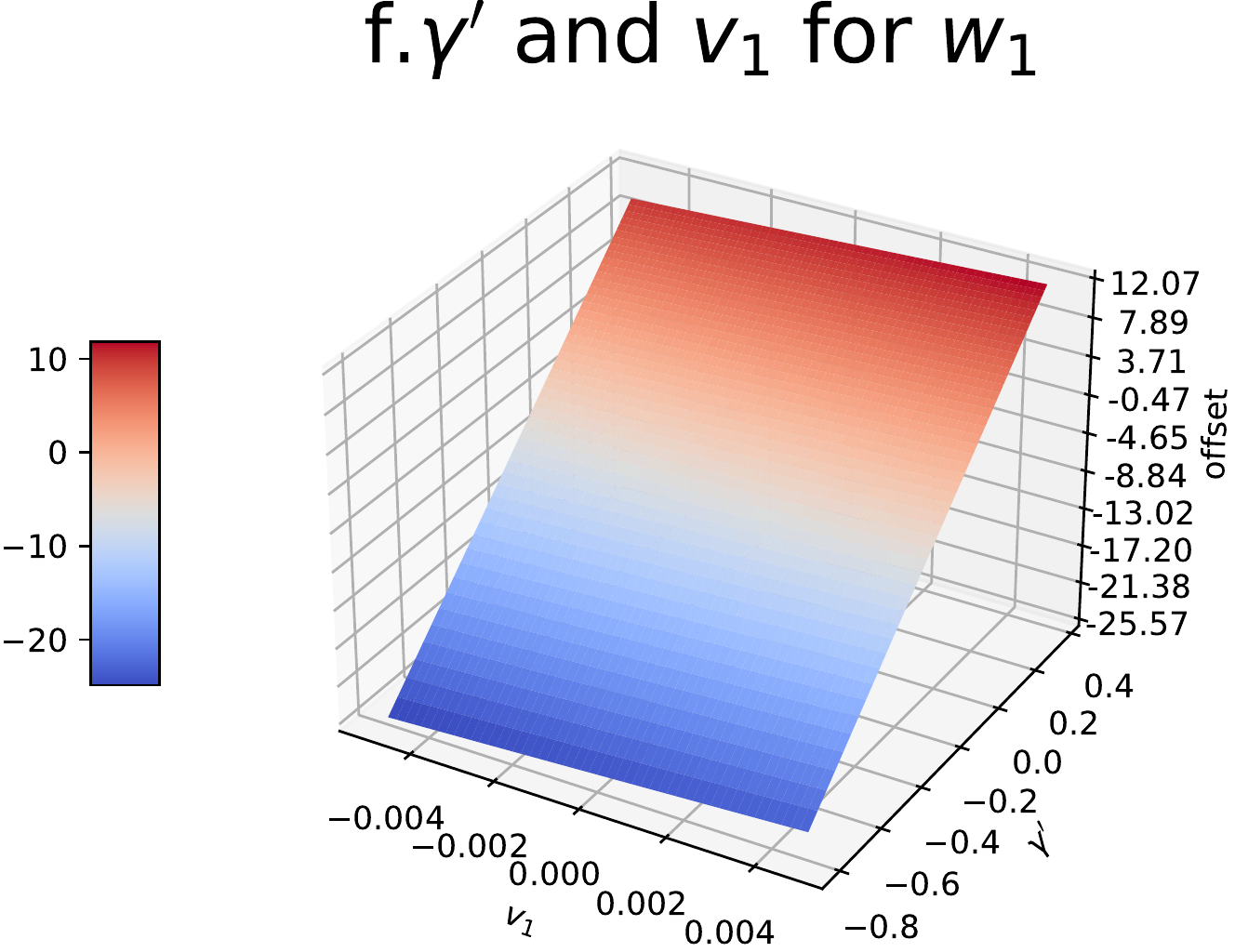}
				%\caption{fig2}
			\end{subfigure}
		\centering
			\begin{subfigure}{0.24\linewidth}
				\centering
				\includegraphics[width=1\linewidth]{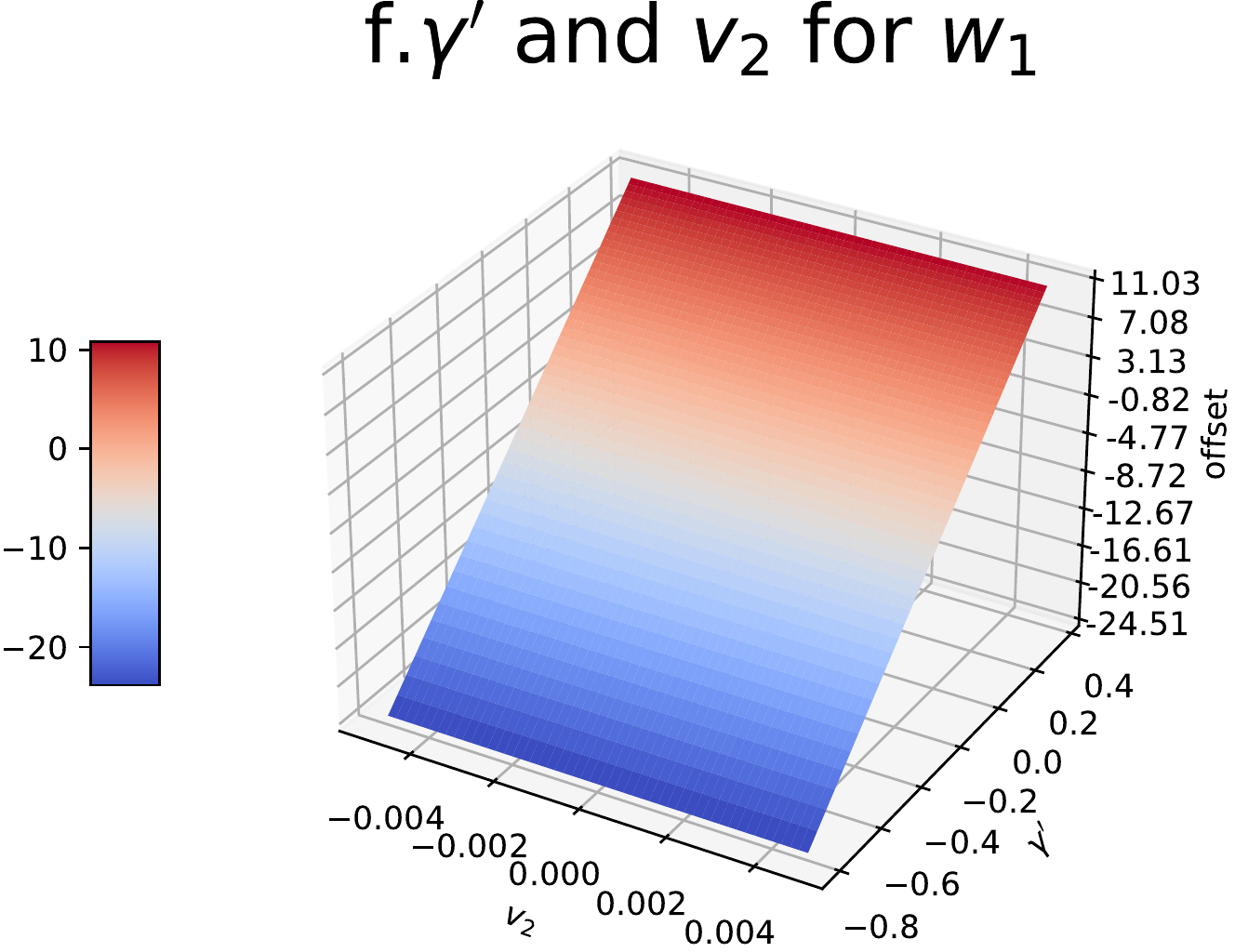}
				%\caption{fig2}
			\end{subfigure}
		\caption{Analysis of the parameter influence to warping function. We use the offsets~(pixels) to measure the influence of $x^\prime$ and other parameters. }
		\label{fig:warp-analysis}
		%		\vspace{-0.2in}
	\end{figure*}

	There is little difference in estimating the two-dimension subgroup of $\mathbf{b}$ and predicting the translation. 
	Given the object center as the origin, all transformations generated by the parameters of $\mathfrak{sl}(3)$ do not change the object's center. 
	Unfortunately, this property does not hold for the warped image. 
	The transformation of $\mathbf{b}$ in the warped image is different from the transformation in the original image. 
	% 	Assuming we have estimated the translation well, 
	To analyze the influence of each parameter on the warping function,
	we draw the center offset of the warped image. One argument is the parameter $\eta_1$ of the warping function, and the other argument is the other parameter $\eta_2$ may influence the translation.
	Fig.~\ref{fig:warp-analysis} shows the example for warping function $w_1$. Fig.~\ref{fig:warp-analysis} (a,b,c,d) demonstrate the $k_x^\prime$,$k_2^\prime$,$\nu_1^\prime$,$\nu_2^\prime$ effect on the warped image center offsets compared with $\theta$ about warp function $w_1$.
	Fig.~\ref{fig:warp-analysis} (e,f,g,h) depict the $k_x^\prime$,$k_2^\prime$,$\nu_1^\prime$,$\nu_2^\prime$ effect on the warped image center in contrast to $\gamma^\prime$ for warp function $w_1$. 
	We find the parameters of $w_1$ dominate the translation of the warped image center. This means that we can estimate $\gamma^\prime$ and $\theta$ in the warped image $g_{w1}$ even with other existing transformations. The same conclusion is valid for other warp functions. 
	
	% 	Note that if there is only the transformation caused by the parameters $\mathbf{x}^\prime $ of the warp function. 
	%%%%---zxr----- delete this, it is not clear
	% {\bf WHAT DO YOU MEAN? In addition, we find the conclusion holds for not only the center point of the warped image but also most of the points offsets around the center decided by the parameters of the warp function. This may be due to the nonlinear and uneven changes for other parameters in the warped image. Thus, the influence on the center offsets does not have a certain direction. The warped image offsets are proportional to $\mathbf{x}^\prime$. Moreover, we achieve a weak equivariant architecture to SL(3) if each entire image offset is dominated by each element of $\mathbf{x}^\prime$.}
	% Besides, enforcing the network to learn the different warped states of the same image may guide the networks to achieve the capability of invariant features about { \bf the image pattern or object identity with transformations. WHAT DO YOU MEAN?}
	%%%%---zxr----- 

	\section{Implementation Details}\label{sec:implementation_details}
	% 	In order to apply WCN in two different tasks, 
	We design two separate architectures with our proposed WCN for classification and planar object tracking, respectively. In this section, we introduce the implementation details of the two tasks with two datasets MNIST-Proj and POT accordingly. 
	
	\paragraph{Classification}
    Two backbone networks are used for the classification task in MNIST-Proj. 
	The first one is a modified LeNet-5~\cite{LeNet}. As described in Table.~\ref{tab:mnist-proj-lenet5}, the localization stage is used to predict the pseudo-translation of the handwritten digits on a warped image compared to the implicit upright digits. 
	Then, we resample the image according to the parameters and concatenate it with the original image as the input for the classification stage. 
	To further examine the capability of our method in the classification task, we implement another classifier to demonstrate the results with a deeper backbone ResNet-18~\cite{ResNet}. 
	As listed in Table~\ref{tab:mnist-proj-res18}, we first use the ResNet-18 to extract the feature, then estimate the transformation parameters $\mathbf{b}^\prime$ with several warp functions.
	Its localization network is the same as the Localization stage in Table~\ref{tab:mnist-proj-lenet5} except that the input size is different.
	According to the estimated $\mathbf{\hat{b}}^\prime$, the resampled image, and the original image are concatenated as the input of another ResNet-18 that uses a two-layer classifier to predict the class of the digits. 

    \paragraph{Planar Object Tracking}
	For the planar object tracking, we treat HDN~\cite{HDN} as our baseline method, which has two warp functions to predict $\mathbf{b}$. 
%	in addition to $[t_1,t_2,\gamma, \theta]$. 
	Besides, the perspective changes are small in the feature map. 
	We thereby estimate $p_1$ and $p_2$ directly on the warped image according to the $w_4$ rather than using the correlation.
	The structure is similar to the homography estimator in HDN, yet we only estimate $\nu_1$ and $\nu_2$ directly.

    \paragraph{Homography Estimation}
    For the homography estimation task, we simply apply the same tracking procedure for estimation. 
    In the testing dataset of SCOCO, transformation is conducted by Eq. (8) in the paper.
	For middle augmentation, we set $\theta\in[0.6\mathrm{rad}, 0.6\mathrm{rad}]$, $\gamma\in[0.7, 1.3]$, $k_1\in[-0.2, 0.2]$, $k_2\in[-0.15, 0.15]$, $\nu_1\in[-0.0001, 0.0001]$ , and $\nu_2\in[-0.0001, 0.0001]$. 
	Large augmentation is with  $\theta\in[0.8\mathrm{rad}, 0.8\mathrm{rad}]$, $\gamma\in[0.7, 1.3]$, $k_1\in[-0.3, 0.3]$, $k_2\in[-0.2, 0.2]$, $\nu_1\in[-0.001, 0.001]$, and $\nu_2\in[-0.001, 0.001]$.

	\subsection{Training}\label{subsec:appendix_training}
	Existing datasets lack the annotations of transformation parameters. 
	Even with those provided, they need to be converted to $\mathbf{b}$ with matrix decomposition, which is not easy.
	Therefore, we augment the possible transformations according to Eq.~(8) in the main paper and transform the images as the training data with the randomly sampled parameters. 
	We augment the dataset MNIST~\cite{Mnist} for classification and GOT-10K~\cite{GOT10k} and COCO-14~\cite{COCO} for planar object tracking during the training, respectively.
	
	For MINST, the model is trained firstly with the supervision on the predicted $\hat{\mathbf{b}}$ and predicted class for $N_e=100$ epochs, which is retrained with only classification loss for $N_e$ in MNIST-Proj. 
	For transformation loss, $\mathcal{L}_T = \mathcal{L}_{sr} + \lambda_1  (\mathcal{L}_t + \mathcal{L}_{k_1} + \mathcal{L}_{k_2}+ \mathcal{L}_{\nu})$, where $\mathcal{L}_{sc}$ is the loss of $(b_3,b_4)$.
	$\mathcal{L}_{t}$ is the loss of translation, and $\mathcal{L}_{k_1}$ is the loss of $b_5$ and $-b_5$.
	$\mathcal{L}_{k_2}$ is the loss of $b_6$, and $\mathcal{L}_\nu$ is the loss of $b_7$ and $b_8$.
	All transformation penalties make use of the robust loss function (i.e., smooth L1) defined in \cite{FastRCNN}. As for POT, we employ the same classification and offset loss as HDN~\cite{HDN} for the newly added parameters in $\mathbf{b}$.
	
% 	The classification networks for MNIST-Proj are trained on a single GPU {\bf 3090???}. 
	For MNIST-Proj classification task, we adopt Adam~\cite{adam} as the optimizer, where the batch size is set to 128. 
	The learning rate starts from 0.001 and decays by a multiplicative factor of 0.95 with an exponential learning scheduler.  Similar to HDN, we trained the whole network 	for planar object tracking on GOT-10k~\cite{GOT10k} and COCO14~\cite{COCO} for 30 epochs with 1 epoch warming up. The batch size is set to $28\times 4$. Our model is trained in an end-to-end manner for 18 hours in our experimental settings. 

    For S-COCO-Proj homography estimation, we use the same training and testing procedure as in the POT, except that we remove the GOT-10k~\cite{GOT10k} from the training datasets. All the methods in the leaderboard are trained with the same augmented dataset with middle augmentation and mask the corner area with a circle mask with the radius of 60 pixels.
    
    \begin{small}
	\begin{table}[htbp]
		% \vspace{-0.05in}
        \setlength{\tabcolsep}{1pt}
        \renewcommand{\arraystretch}{1} % Default value: 1
		\caption{Network details for MNIST-Proj. Conv~(1,8,7) denotes a convolution layer with input channel=1, output channel=8, kernel size=7. MaxPool~(2,2) represents the max-pooling layer with window=2, and stride=2. Linear~(90,32) represents the fully connected layer with input size=90 and output size=32. C is the total number of channels for the output.}
		\label{tab:mnist-proj-lenet5}
		\centering
		\begin{tabular}{lll}
			\toprule
			%			\multicolumn{2}{c}{Part}                   \\
			\cmidrule(r){1-3}
			Stages           & Operator   & Output \\
			\midrule
			&     Conv2d~(1,8,7)        & C$\times$8$\times$22$\times$22   \\
			&     MaxPool~(2,2), ReLU  & C$\times$8$\times$11$\times$11    \\
			Localization &     Conv2d~(8,10,5)  & C$\times$10$\times$7$\times$7    \\\
			&     MaxPool~(2,2), ReLU  & C$\times$10$\times$3$\times$3   \\
			&     Linear~(90,32), ReLU  &  C$\times$32    \\
			&     Linear~(32,2)        & C$\times$2     \\
			\hline
			&	 Conv2d~(2,10,5)        & C$\times$10$\times$24$\times$24    \\
			&     MaxPool~(2,2), ReLU  & C$\times$10$\times$12$\times$12   \\
			Classification    &     Conv2d~(10,20,5)  &  C$\times$20$\times$8$\times$8    \\\
			&     MaxPool~(2,2), ReLU, Dropout  &   C$\times$20$\times$4$\times$4    \\
			&     Linear~(90,32), ReLU  &  C$\times$50    \\
			&     Linear~(50,10)        & C$\times$10    \\
			
			% 			Naive(wo.aug)     & LeNet5$^*$    & 59.37 ($\pm0.99$)     \\
			% 			Navie             & ResNet18  & 0.87 ($\pm$0.13)  \\			
			% 			STN(\cite{STN})  & LeNet5$^*$ & 4.00 ($\pm$0.35)\\
			% 			STN               & ResNet18  & 0.79($\pm$0.07)  \\
			% 			Ours              & LeNet5$^*$  & 3.05 ($\pm$0.33)  \\
			% 			Ours              & ResNet18  & 0.69 ($\pm$0.09)  \\
			\bottomrule
		\end{tabular}
		% 		\vspace{-0.2in}
	\end{table}
    \end{small}
	\begin{table}[htbp]
		% \vspace{-0.05in}
          \setlength{\tabcolsep}{1pt}
        \renewcommand{\arraystretch}{1} % Default value: 1
		\caption{Classification Networks details using ResNet18}
		\label{tab:mnist-proj-res18}
		\centering
		\begin{tabular}{lll}
			\toprule
			%			\multicolumn{2}{c}{Part}                   \\
			\cmidrule(r){1-3}
			Stages           & Operator   & Output \\
			\midrule
			Classification	&	 Linear (3136,128), Norm,ReLU        & C$\times$128    \\
			&   Linear (128,10), LogSoftmax&C$\times$10    \\
			\bottomrule
		\end{tabular}
		% 		\vspace{-0.2in}
	\end{table}
		\begin{figure*}[t]
		\centering
        \begin{subfigure}{0.4\linewidth}
            \centering
            \includegraphics[width=1\linewidth]{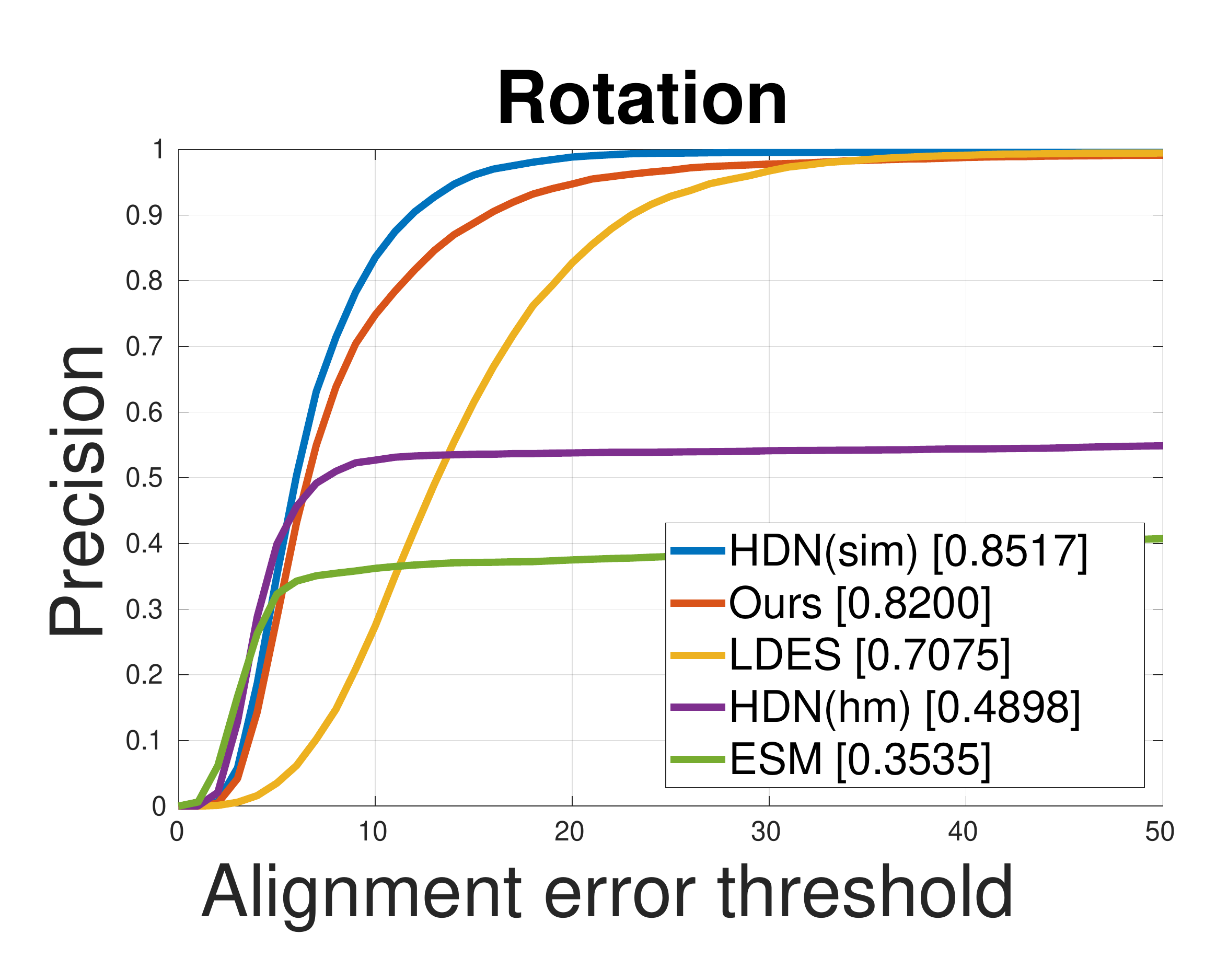}
            %\caption{fig1}
        \end{subfigure}%
        \begin{subfigure}{0.4\linewidth}
            \centering
            \includegraphics[width=1\linewidth]{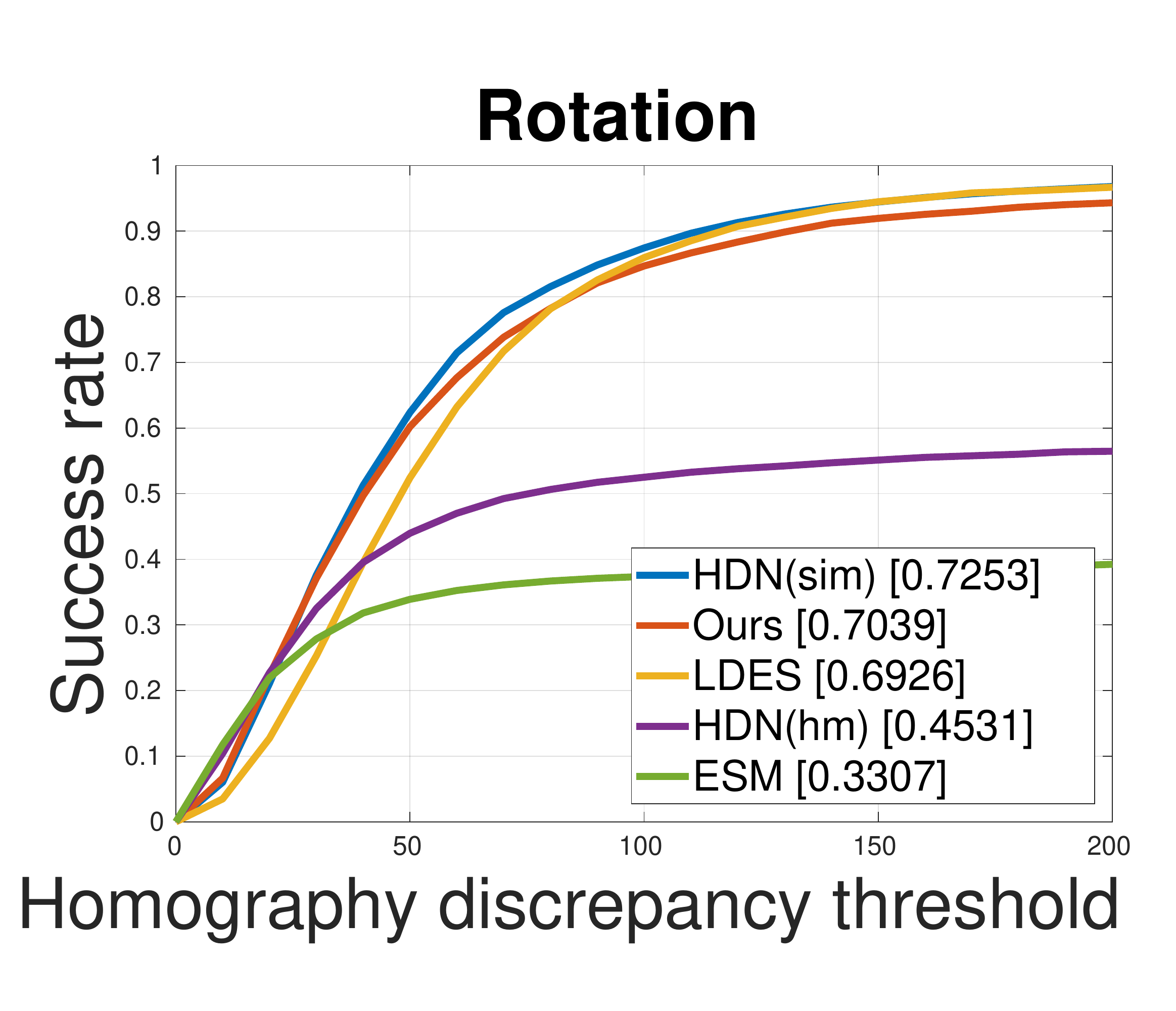}
            % 			\includegraphics[width=1\linewidth]{figs/experiment/All-seq-precision.pdf}
            %				\caption{fig2}
        \end{subfigure}

        \begin{subfigure}{0.4\linewidth}
            \centering
            \includegraphics[width=1\linewidth]{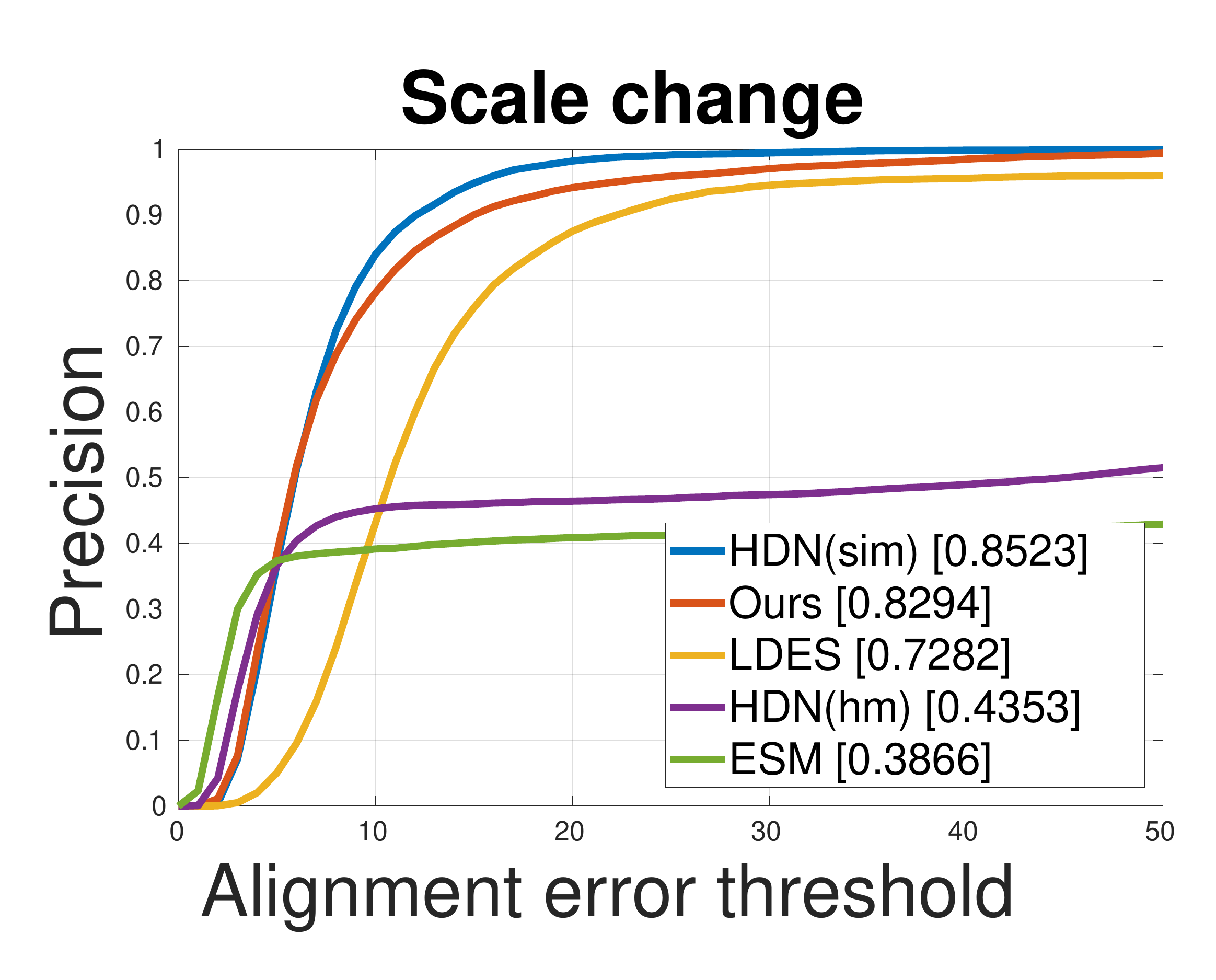}
            % 			\includegraphics[width=1\linewidth]{figs/experiment/All-seq-precision.pdf}
            %				\caption{fig2}
        \end{subfigure}
        \begin{subfigure}{0.4\linewidth}
            \centering
            \includegraphics[width=1\linewidth]{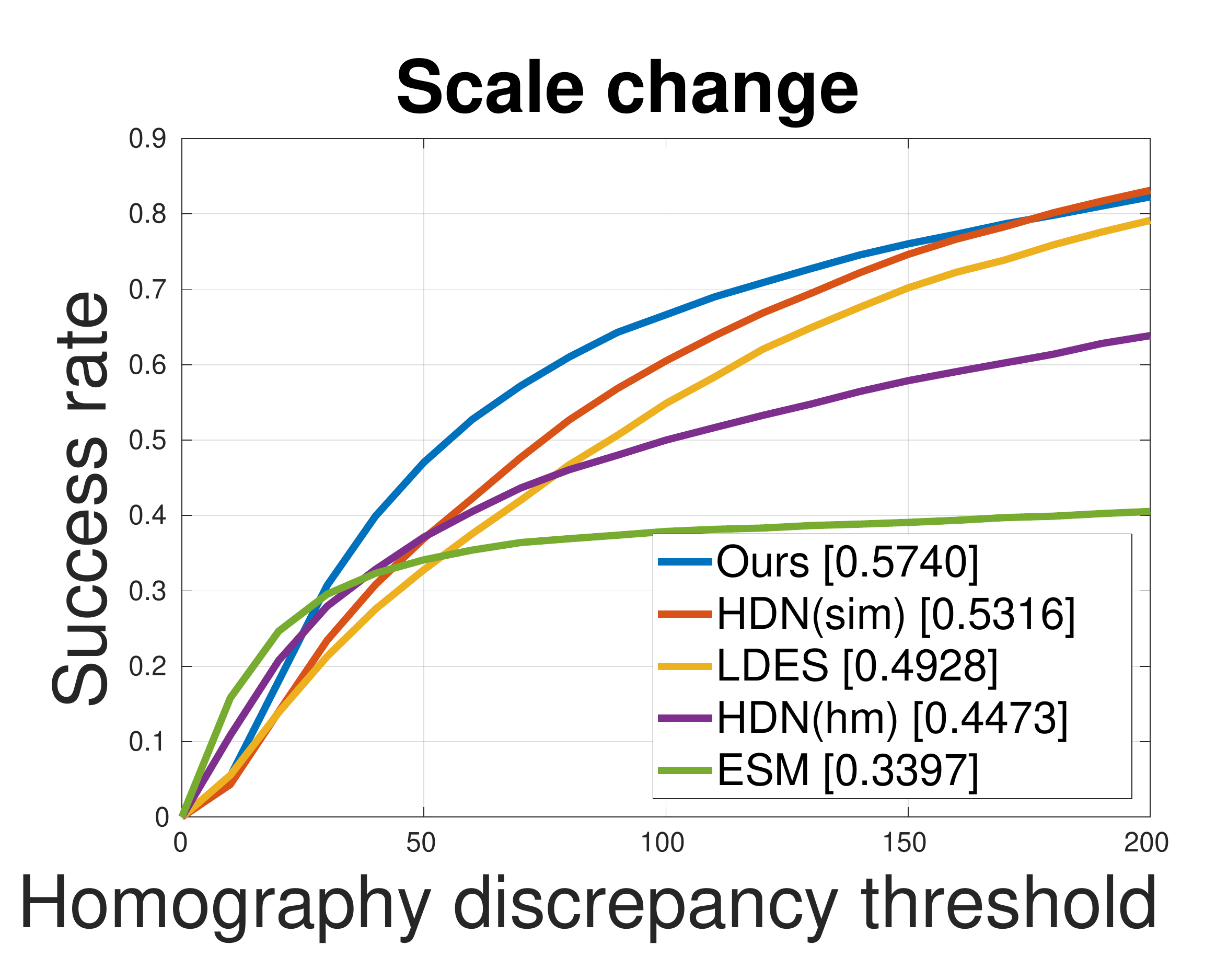}
            % 			\includegraphics[width=1\linewidth]{figs/experiment/All-seq-precision.pdf}
            %				\caption{fig2}
        \end{subfigure}
		\centering
		\caption{ Results on rotation and scale sequences of POT. }
        % \vspace{-0.1in}

		\label{fig:pot_compare_supp}
	\end{figure*}
 
	\section{Experiments}\label{sec:appendix_experiments}
	
	\subsection{Experimental Setup}\label{subsec:appendix_experimental_setup}
	% or time epochs datasets. etc.
	We conducted all experiments on a PC with an intel E5-2678-v3 processor~(2.5GHz), 32GB RAM, and an Nvidia GTX 2080Ti GPU. 
	The proposed method is implemented in Pytorch. 
	
	For MNIST-Proj, the size of the input image is $28\times28$. 
	For the hyperparameters of WCN in training, we set $\lambda=2$, $\lambda_1=20$, $\gamma \in [1/1.4, 1.4]$, $\theta \in [-1.5, 1.5]$, $t\in[-28/8,28/8]$, $k_1\in[-1.3, 1.3]$, $k_2 \in [-0.03,0.03]$, and $\nu_1,\nu_2 \in [-0.02,0.02]$.  Due to the numbers 6 and 9 being identical with the rotation even from humans, we remove the number 9 from MNIST-Proj.   
	
	For POT, the size of input template $T$ for our networks is $127\times 127$, while search image $I$ has the size of $255\times 255$ to deal with the large homography changes. 
	All the hyper-parameters are set empirically, and we do not use any re-initialization and failure detection scheme. 
	For the hyper-parameters of HDN in training, we set 	
	$\gamma \in [1/1.38, 1.38]$, $\theta \in [-0.7, 0.7]$, $t\in[-32,32]$, $k_1\in[-0.1, 0.1]$, $k_2 \in [-0.015,0.015]$ and $\nu_1,\nu_2 \in [-0.0015,0.0015]$.
	
	%\subsection{More Results on Mnist}
	% 
	%\begin{table}[t]
	%	\vspace{-0.05in}
	%	\caption{MNIST-Proj results.}
	%	\label{tab:mnist-comp}
	%	\centering
	%	\begin{tabular}{ll}
	%		\toprule
	%		%			\multicolumn{2}{c}{Part}                   \\
	%		\cmidrule(r){1-2}
	%		Methods              & Error ($\%$) \\
	%		\midrule
	%		% 			Naive(\cite{LeNet})     & 11.48 ($\pm1.42$)     \\
	%		L-conv (\cite{LConv})   & 19.16 ($\pm$1.84)  \\
	%		homConv (\cite{Macdonald2021EnablingEF})    & 14.72 ($\pm$0.72)  \\
	%		PDO-econv (\cite{shen2020pdo}) & 1.66 ($\pm$0.16)  \\
	%		LieConv(\cite{finzi2020generalizing})    & 2.7 ($\pm$0.74)  \\
	%		PTN (\cite{PTN}) & 2.45($\pm$0.66)  \\
	%		STN (\cite{STN}) & 0.79 ($\pm$0.07)  \\
	%		Ours                & \textbf{0.69} ($\pm$0.09)  \\
	%		\bottomrule
	%	\end{tabular}
	%	\vspace{-0.2in}
	%\end{table}

	Many parameters may influence the experimental results. We investigate the influence of the sampling circle radius. We fix it to be $n/2$, which is the half length of the side in the default setting.
	Theoretically, it is enough as long as the field covers the region of the original image. 
	Actually, we tested 5 different radius $(0.5\times n/2, 0.75\times n/2, n/2, 1.25\times n/2, 1.5\times n/2)$
	The results are quite similar, and the errors are within $0.1\%$.
	%For other evaluation, we will add more applications and experiments on heuristics in the next step journal paper.
	
	\subsection{More Results on POT}
	Apart from the perspective distortion in the main paper, we provide more results on other simple transformations, e.g. rotation and scale changes. 
	Fig.~\ref{fig:pot_compare_supp} shows the precision and success rate of these two transformations. 
	HDN~\cite{HDN} has a similar structure for similarity estimation with two same warp function, which is similar to our presented WCN. 
	Thereby, we have similar results on rotation and scale sequences. 
	Our performance is inferior to HDN~\cite{HDN} when the error threshold is small. 
	When there are either rotation or scale changes for the object, our WCN estimates all eight parameters rather than similarity transformation compared to HDN. This may bring more estimation errors because the estimation is not accurate as explained in Sec.~\ref{sec:warp-func}.
	Furthermore, HDN uses more training data than WCN does.
	When the error threshold is large, their precision and success rate are close. 

\section{Limitation and Future Work}\label{sec:limitation}
	% \vspace{-2mm}
	%	1. not theoretical invariant, can not explain why... may relate to the motion distortion. Because the only translation, others random. not the same direction.
	%	2. occlusion problem future work.
	%	3. accuracy. Robustness.
	% 	It may be related to the relative transformation degree and the distribution that influences the object's centroid. and the process of methods determines the estimation results is coarse,
	Whereas our proposed method is theoretically equivariant to the mentioned several groups, it is hard to theoretically prove why CNNs can robustly estimate a series of transformation parameters sequentially. 
	Besides, a known problem of WCN is that the estimated offsets may be inaccurate due to the influence of other parameters in addition to $\mathbf{b}^\prime$ and the small feature map size and error produced in different $M_i$. 
	The warped image may thereby bring the error of the previously predicted parameters. 
	Moreover, the sampling density also affects the estimation accuracy. Although no additional parameters will be added to the networks, the number of resampling times is the same as the number of inference times through the networks. It may be time-consuming if there are too many warping functions in a group.
	In our experiments, the proposed method is not robust to the challenging scenarios like partial occlusions and heavy blur, which can be solved by predicting either an extra occlusion map or a blur kernel. Furthermore, more applications such as AR, SLAM, recognition, congealing, and image stabilization can be benefited from our proposed method.%This is closely related to 

% 	Therefore, the WCN requires a refinement module if one needs a more accurate transformation.
	
	\section{Proofs}
We define the warp function $w$ for different elements in $\mathbf{b}$, and let $\mathbf{b}^\prime$ be the elements in $\mathbf{b}$ with regard to each warp function $w_i$. 
Although the coordinates of the warped image are proportion to $\mathbf{b}^\prime$, we still need to prove that the group action results on the sampled points $\mathbf{u}^\prime$ in the original image are additive about $\mathbf{b}^\prime$. 
That is,
\begin{align}
	&\quad \mathbf{H}(\Delta \mathbf{b}^\prime) \cdot \mathbf{u}^{\prime T} \\&= \mathbf{H}(\Delta \mathbf{b}^\prime) \cdot w(\mathbf{b}^\prime) \\&= w(\mathbf{b}^\prime+\Delta \mathbf{b}^\prime)
	\label{eq:warp-function-satis}
\end{align}
where $\mathbf{H}$ can be viewed as a function of $\mathbf{x}^\prime$, and $\mathbf{x}^\prime$ can be viewed as a function of $\mathbf{b}^\prime$. 
$\mathbf{\Delta b}^\prime$ is the incremental value of $\mathbf{b}^\prime$. 
Therefore, the warped function satisfies Eq.~(6) in the main paper, which makes the convolution equivariant to $\mathbf{b}^\prime$.

%	\subsection{Proof for warp functions}

\noindent\textbf{Scale and Rotation}\\
%	let x be the coordinates of a point in image $I$, The transformed point by $H_s$ is:
As introduced in the main paper, the warp function for scale and rotation is :
\begin{equation}
	%	g_{\gamma,\theta}(\mathbf{\tau}) = 
	w_{1}(b_3, b_4) = (u_1^\prime, u_2^\prime)^T=
	\begin{bmatrix}
		s^{\gamma^\prime} \cos(b_3) \\
		s^{\gamma^\prime} \sin(b_3)
	\end{bmatrix}
	\label{eq:warp_lp_supp}
\end{equation}
We have defined $\gamma = s^{\gamma^\prime} = e^{b_4}$ and $\theta=b_3$, Therefore, the left of Eq.~\ref{eq:warp-function-satis} can be rewritten as below: 
%	\begin{equation}
\begin{align}
	&\quad\mathbf{H}_s((\Delta b_3, \Delta b_4))\cdot \mathbf{u}^{\prime T} = \mathbf{H}_s\cdot w_1 \\&=  	
	\begin{bmatrix} 
		e^{\Delta b_4} \cos(\Delta b_3)u_1^\prime-e^{\Delta b_4} \sin(\Delta b_3) u_2^\prime\\
		e^{\Delta b_4} \sin(\Delta b_3)u_1^\prime +e^{\Delta b_4} \cos(\Delta b_3) u_2^\prime
	\end{bmatrix}\\&= 
	\begin{bmatrix} 
		e^{b_4+\Delta b_4 }\cos(b_3+\Delta b_3 ) \\
		e^{b_4+\Delta b_4 }\sin(b_3+\Delta b_3 ) \\
	\end{bmatrix}
	\label{eq:warp_lp_proof}
	%	\end{equation}
\end{align}
As a result, the warp function $w_1$ supports the equivariance. 
%	Let $\theta(0)$ be the $\theta$ when there is no transformation from $\theta$ on the transformed I, and from the warping function Eq.~\ref{eq:warp_lp} and group action formula Eq.~\ref{eq:action_lp}, one can derive that:  
%	\begin{equation}
%	\theta - \theta(0) = arctan(x'_1, x'_2) - arctan(x_1, x_2) = \theta
%	\end{equation}
%	Let $\gamma(0)$ be the warped $gamma'$ of the transformed I. 
%	\begin{equation}
%	\gamma - \gamma(0) = log(||(x')||_2) - log(||(x)||_2) =  \gamma
%	\end{equation}
%	Hence, after warping the derived theta performs the same as translation which is consistent with Eq.~\ref{eq:warped-conv}

%	\noindent\textbf{Proof. sc}
\noindent\textbf{Aspect Ratio}\\
The warp function for aspect ratio is defined as follows:
\begin{equation}
	% 	\vspace{-0.1in}
	%	g_{k_1^{\prime},1/k_1^{\prime}}(\mathbf{\tau}) = 	
	w_{2}(b_5,-b_5) = 
	\begin{bmatrix}
		s^{k_x^\prime} , s^{k_y^\prime}
	\end{bmatrix}^T,
	\label{eq:warp-dl}
\end{equation}
For $\mathbf{H}_{sc}$, there is only one element.
%We thereby constrain $e^{b_5} = k_1 = k_x = 1/k_y=s^{k_x^\prime}$. 
We thereby let $k_1 = (s^{k_x^\prime}=\exp({k_x^\prime\log s})) = \exp({b_5})$ and $1/k_1 = (s^{-k_x^\prime}=\exp({-k_x^\prime\log s})) = \exp({-b_5})$.
As a result, the left of Eq.~\ref{eq:warp-function-satis} can be rewritten as below
\begin{align}
	&\quad\mathbf{H}_{sc}((\Delta k_x^\prime, \Delta k_y^\prime)) \cdot \mathbf{u}^{\prime T} = \mathbf{H}_{sc} \cdot w_2 \\&= 
	\begin{bmatrix}
		e^{\Delta b_5}\cdot u_1^\prime \\
		e^{-\Delta b_5}\cdot u_2^\prime
	\end{bmatrix}
	=
	\begin{bmatrix}
		e^{\Delta b_5+b_5}  \\
		e^{-(\Delta b_5+b_5)} 
	\end{bmatrix}
	%	
	%		w_{2}(k_x,k_y) = 
	%	\begin{bmatrix}
	%	s^{k_x^\prime} , s^{k_y^\prime}
	%	\end{bmatrix}^T,
	\label{eq:dl-proof}
\end{align}	
Hence, we can prove the equivariance holds with the warp function $w_2$.

\noindent\textbf{Shear}\\
% \textbf{Shear} \\
The warp function for Shear is defined as below:
\begin{equation}
	w_{3}(b_6, b_\epsilon) = 
	\begin{bmatrix}
		b_6 b_\epsilon, & 
		b_\epsilon 
	\end{bmatrix}^T
	\label{eq:warp_sh_supp}
\end{equation}
The left of Eq.~\ref{eq:warp-function-satis} can be rewritten as:
\begin{align}
	&\quad \mathbf{H}_{sh}((\Delta b_6, b_\epsilon)) \cdot \mathbf{u}^{\prime T} = \mathbf{H}_{sh} \cdot w_3\\ &= 
	\begin{bmatrix}
		u_1^\prime + \Delta b_6 u_2^\prime \\
		u_2^\prime	
	\end{bmatrix}
	=
	\begin{bmatrix}
		(b_6+\Delta b_6) b_\epsilon  \\
		b_\epsilon
	\end{bmatrix}
	%	
	%		w_{2}(k_x,k_y) = 
	%	\begin{bmatrix}
	%	s^{k_x^\prime} , s^{k_y^\prime}
	%	\end{bmatrix}^T,
	\label{eq:sh-proof}
\end{align}	
Hence, the equivariance is tenable for $\mathbf{H}_{sh}$ with warp function $w_3$.

% \textbf{Perspective}\\
\noindent\textbf{Perspective}\\
The warp function for perspective can be derived as below:
%	\begin{equation}
%	\mathbf{u}^* = \mathbf{H}_{p} \cdot \mathbf{u}^T =  \mathbf{H}_{p2} \mathbf{H}_{p1}\cdot \mathbf{u}^T =
%	\begin{bmatrix} 
%	u_1/(\nu_1 u_1 + \nu_2 u_2+1) ,& 
%	u_2/ (\nu_1 u_1 + \nu_2 u_2+1)
%	\end{bmatrix}^T
%	\label{eq:perspective-action}
%	\end{equation}
\begin{equation}
	w_{4}(b_7, b_\epsilon) = 
	\begin{bmatrix}
		\frac{1}{b_7}, &
		\frac{b_\epsilon}{b_7}
	\end{bmatrix}^T,  
	w_{5}(b_\epsilon, b_8) = 
	\begin{bmatrix}
		\frac{b_\epsilon}{b_8}, &
		\frac{1}{b_8}
	\end{bmatrix}^T
	\label{eq:warp_ps_supp}
\end{equation}
We have defined the $\nu_1 = b_7$ and $\nu_2 = b_8$. Thus, the left of Eq.~\ref{eq:warp-function-satis} with regard to $\nu_1$ or $\nu_2$ is rewritten as follows:

\begin{align}
	&\quad\mathbf{H}_{p1}((\Delta b_7, b_\epsilon)) \cdot \mathbf{u}^{\prime T} = \mathbf{H}_{p1} \cdot w_4 \\&= 
	\begin{bmatrix}
		\frac{u_1^\prime}{u_1^\prime \Delta b_7 + 1} \\
		\frac{u_2^\prime}{u_1^\prime \Delta b_7 + 1} 
	\end{bmatrix}
	=
	\begin{bmatrix}
		\frac{1}{\Delta b_7 + b_7} \\
		\frac{b_\epsilon}{\Delta b_7+ b_7}
	\end{bmatrix}
	%	
	%		w_{2}(k_x,k_y) = 
	%	\begin{bmatrix}
	%	s^{k_x^\prime} , s^{k_y^\prime}
	%	\end{bmatrix}^T,
	\label{eq:ps1-proof}
\end{align}	

\begin{align}
	&\quad\mathbf{H}_{p2}((b_\epsilon, \Delta b_8)) \cdot \mathbf{u}^{\prime T} = \mathbf{H}_{p2} \cdot w_5 \\&= 
	\begin{bmatrix}
		\frac{u_1^\prime}{u_2^\prime \Delta b_8 + 1} \\
		\frac{u_2^\prime}{u_2^\prime \Delta b_8 + 1} 
	\end{bmatrix}
	=
	\begin{bmatrix}
		\frac{b_\epsilon}{\Delta b_8 + b_8} \\
		\frac{1}{\Delta b_8+ b_8}
	\end{bmatrix}
	\label{eq:ps2-proof}
\end{align}	
Hence, the equivariance holds for two perspective groups with warp function $w_4$ and $w_5$, respectively.
%%%%%%%%% REFERENCES
{\small
\bibliographystyle{ieee_fullname}
\bibliography{egbib}
}

\end{document}